\documentclass[10pt,twocolumn,letterpaper]{article}

\usepackage[pagenumbers]{cvpr}      % To produce the REVIEW version
\usepackage{mathtools}
\usepackage{cirl}
\usepackage[dvipsnames]{xcolor}

\definecolor{cvprblue}{rgb}{0.21,0.49,0.74}
\usepackage[pagebackref,breaklinks,colorlinks,citecolor=cvprblue]{hyperref}

\def\paperID{249} % *** Enter the Paper ID here
\def\confName{3DV\xspace}
\def\confYear{2026\xspace}

\title{SmokeSeer: 3D Gaussian Splatting for Smoke Removal and Scene Reconstruction}

\author{Neham Jain, Andrew Jong, Sebastian Scherer, Ioannis Gkioulekas\\
Carnegie Mellon University\\
Pittsburgh, PA\\
{\tt\small nhjain@andrew.cmu.edu, ajong@andrew.cmu.edu, basti@andrew.cmu.edu, igkioule@cs.cmu.edu}}

\begin{document}
%\maketitle

\twocolumn[{%
\renewcommand\twocolumn[1][]{#1}%
\maketitle
\includegraphics[width=\linewidth]{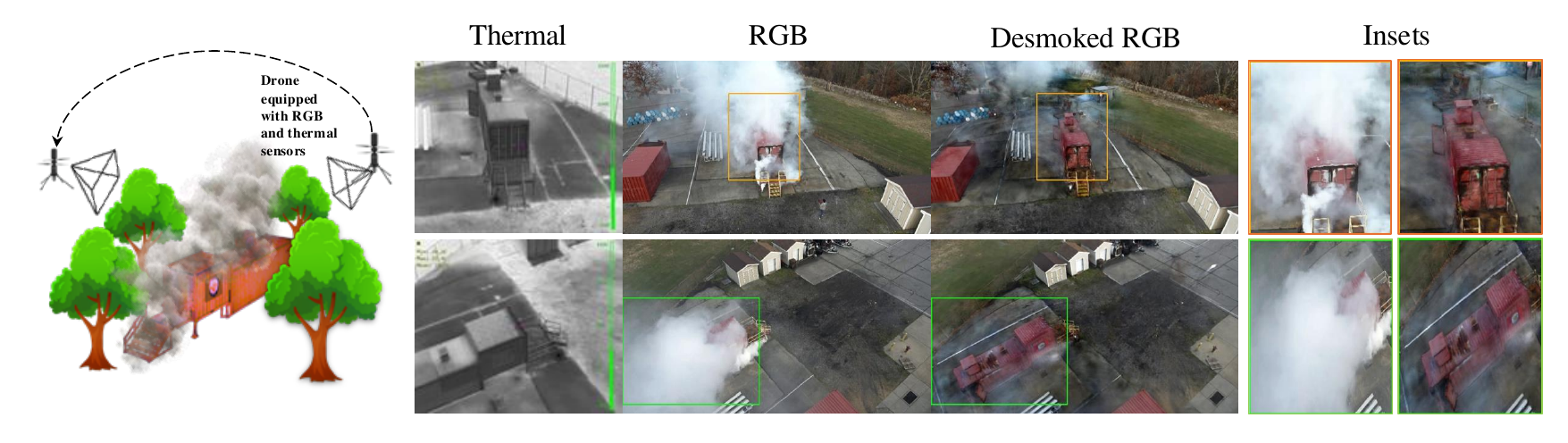}
\captionof{figure}{Our method utilizes RGB and thermal images from a drone-mounted sensor to perform simultaneous 3D scene reconstruction and smoke removal
using an inverse rendering approach within the 3D Gaussian splatting framework. Insets highlight the effectiveness of our approach in revealing occluded structures.
\vspace{0.3em}}
\label{fig:teaser}
}]

\begin{abstract}
    Smoke in real-world scenes can severely degrade image quality and hamper visibility. 
    Recent image restoration methods either rely on data-driven priors that are susceptible to hallucinations, or are limited to static low-density smoke. We introduce \emph{SmokeSeer}, a method for simultaneous 3D scene reconstruction and smoke removal from multi-view video sequences. Our method uses thermal and RGB images, 
    leveraging the reduced scattering in thermal images to see through smoke.
    We build upon 3D Gaussian splatting to fuse information from the two image modalities, 
    and decompose the scene into smoke and non-smoke components.
    Unlike prior work, SmokeSeer handles a broad range of smoke densities and adapts to temporally varying smoke. 
    We validate our method on synthetic data and a new real-world smoke dataset with RGB and thermal images. 
    We provide an open-source implementation and data on the project website.\footnote{\urlstyle{same}\url{https://imaging.cs.cmu.edu/smokeseer}}
\end{abstract}
    
\section{Introduction}
\label{sec:intro}

Reliable visual perception is essential for safety-critical applications such as search and rescue, robot navigation, 
and industrial inspection. The ability to accurately perceive and reconstruct 3D environments is particularly vital, 
as it enables precise spatial reasoning and path planning in complex scenarios. 
For example, firefighters navigating through burning buildings increasingly depend on vision-based systems to 
maintain situational awareness. However, dense smoke severely compromises these systems, 
obscuring vital environment details and increasing operational risks. 
Developing technologies that enable these systems to ``see through smoke'' is therefore critical for 
enhancing both safety and operational effectiveness in these hazardous environments.

Though several approaches have targeted the problem of enhancing visibility through scattering media,
significant limitations remain. Learning-based approaches that map hazy to clear images require extensive paired 
datasets and typically process individual frames, thereby ignoring valuable multi-view constraints.
Closer to our work, neural rendering approaches such as ScatterNeRF \cite{Ramazzina_2023_ICCV} 
and DehazeNeRF \cite{Chen2024dehazenerf} incorporate physical light transport models and operate on multi-view RGB data. 
However, all these approaches primarily address static haze removal and are ill-equipped to handle dense, temporally evolving smoke.

We build an end-to-end system that performs joint 3D scene reconstruction and smoke removal in the presence of dense, temporally evolving smoke. 
Our method uses images from RGB and thermal cameras, and is effective on real-world smoke data (Figure~\ref{fig:teaser}).
We build upon 3D Gaussian splatting (3DGS) \cite{kerbl3Dgaussians} and decompose a smoke-filled scene into 
two sets of Gaussians: one representing the smoke part, and another representing the non-smoke part of the scene, 
which we refer to as the surface Gaussians. 
This decomposition allows us to render only the surface Gaussians to visualize the scene without smoke.

Performing this decomposition using only RGB images is challenging due to the visual ambiguity between 
light-reflecting surfaces and light-scattering smoke particles.
To address this challenge, we leverage thermal cameras that capture long-wavelength infrared radiation, 
which is substantially less affected by scattering in smoke than visible light. This property enables thermal sensors 
to preserve critical spatial information even in dense smoke conditions. However, thermal images are 
low-resolution, have low contrast, and lack the texture details crucial for object recognition and scene understanding. 
Our method overcomes this limitation through a joint optimization strategy that fuses the robust spatial cues 
from thermal data with the rich texture information provided by RGB imagery.

% To effectively model temporally evolving smoke—a characteristic absent in prior work—
% we introduce a deformation field that captures the dynamic nature of smoke particles. 
% This allows our system to maintain consistent scene reconstruction despite the continuously 
% changing smoke patterns, a crucial capability for practical deployment in emergency response scenarios.

%Talk a bit on how we model temporally evolving smoke and why doing that is a good thing 

To effectively leverage the complementary strengths of both modalities, we propose a three-stage approach for smoke removal and 3D scene reconstruction. 
In the first stage, we leverage advances in 3D foundation models \cite{Wang_2024_CVPR} to estimate RGB-thermal poses in the same coordinate frame. 
In the second stage, we learn the scene's geometry exclusively from thermal images, 
leveraging their robustness in capturing spatial information even in the presence of dense smoke. 
In the third stage, we use both RGB and thermal images to optimize two sets of Gaussians, for the smoke and the scene's surfaces. 
For the surface Gaussians, we rely on initialization from the output of the second stage. 
For the smoke Gaussians, we use a deformation field to model the temporal variation of smoke, 
and enforce handcrafted priors based on physical properties of smoke. These choices help ensure that, after optimization, 
smoke Gaussians exclusively capture the scene smoke, whereas surface Gaussians accurately represent the underlying scene structure.

Unlike prior learning-based dehazing methods, ours does not directly rely on image-to-image learned priors 
and instead formulates smoke removal as an inverse rendering problem within the 3DGS framework. 
To the best of our knowledge, this is the first work that jointly uses RGB and thermal images for smoke removal and 3D reconstruction. 

Our experiments show state-of-the-art results on both simulated and real-world datasets---collected in partnership with our county's fire department using a field operational drone---for smoke removal and novel view synthesis. Our code and data are publicly available on the project website, to ensure reproducibility and facilitate follow-up research.

\section{Related Work}

\label{sec:related_work}

\subsection{Image-based methods for haze removal}

\paragraph{Traditional methods.}
Koschmieder \cite{koschmieder1924theorie} developed an atmospheric scattering model that describes image formation under haze as a combination of 
direct attenuation and airlight. This model is a simplification of the more general radiative 
transfer equation (RTE) \cite{chandrasekhar1960radiative}, which describes the propagation of light through a medium with scattering and absorption.
Though widely used in dehazing methods, the Koschmieder model assumes homogeneous static media, limiting its effectiveness for heterogeneous, 
dynamic smoke conditions.

Early image restoration approaches relied on handcrafted priors to estimate physical parameters in the Koschmieder model. 
He et al. \cite{he2011single} tried to estimate the attenuation map by leveraging the observation that in most local patches of haze-free images, at least one color channel has very low intensity.
Zhu et al. \cite{zhu2015fast} proposed the color attenuation prior, modeling the depth of the scene through the difference between brightness and saturation.
Berman et al. \cite{berman2016non} developed a non-local method based on the observation that colors in haze-free images form tight clusters in RGB space.
Though effective for thin homogeneous haze, these methods fail in dense smoke scenarios, for which their priors are ill-suited.

\paragraph{Learning-based methods.}
Some recent methods map hazy to clear images without explicit parameter estimation.
Examples include MSRL-DehazeNet~\cite{liu2019multi}, 
collaborative inference frameworks for dense haze in remote sensing~\cite{densehaze2023dynamic},
and saliency-guided mechanisms for UAV imagery~\cite{uav2024saliency}.

Transformer-based architectures have recently shown promising results for dehazing. 
Zamir et al. \cite{li2022you} proposed Restormer, an efficient transformer for high-resolution image restoration including dehazing.
Guo et al. \cite{guo2022image} introduced a hybrid CNN-transformer architecture that combines local and global feature extraction.
Despite these advances, most learning-based methods process individual frames independently, ignoring valuable temporal and multi-view information that could enhance smoke removal performance.

Specific to smoke removal, Salazar-Colores et al. \cite{fujita2022smoke} developed an image-to-image translation approach guided by an embedded dark channel for desmoking laparoscopy surgery images.
However, this and other similar methods typically require paired training data (smoke versus smoke-free), which is challenging to obtain in real-world scenarios, especially for temporally varying smoke.

\subsection{Neural representations for participating media}

Neural radiance Fields (NeRF)~\cite{mildenhall2020nerf} have revolutionized scene representation using continuous volumetric functions. 
Several works have extended NeRF to handle participating media such as smoke and haze.
ScatterNeRF~\cite{Ramazzina_2023_ICCV} incorporates the Koschmieder model into the NeRF framework, but remains limited to homogeneous haze conditions. DehazeNeRF~\cite{Chen2024dehazenerf}
can handle heterogeneous media but not dynamic smoke. 
These methods have primarily focused on static haze removal and do not address the more challenging problem of temporally varying smoke---our focus. 

3D Gaussian splatting (3DGS)~\cite{kerbl3Dgaussians} is an efficient alternative to NeRF through 
scene representation using 3D Gaussians, enabling real-time rendering.
Dynamic 3DGS \cite{luiten2023dynamic} extends this framework to dynamic scenes, but does not specifically address participating media. 

Lastly, recent approaches such as ThermalNeRF~\cite{lin2024thermalnerf} and ThermalGaussian splatting~\cite{lu2024thermalgaussian} 
incorporate thermal imaging into neural rendering frameworks but do not tackle the problem of imaging through smoke.

\subsection{Multi-modal sensing}

Multi-modal sensing has emerged as a promising direction for robust perception in challenging environments.
Thermal imaging, which captures long-wavelength infrared radiation, is less affected by smoke and haze compared to RGB cameras \cite{amin2022thermal}.
Hwang et al. \cite{hwang2015multispectral} demonstrated the effectiveness of fusing RGB and thermal information for object detection in adverse weather.
Li et al. \cite{chen2021all} proposed an RGB-thermal object tracking benchmark demonstrating the value of thermal information for robust perception.

% For 3D reconstruction, Khattak et al. \cite{khattak2023thermal} used thermal-inertial odometry for robust localization in smoke-filled environments.
% Bijelic et al. \cite{bijelic2020seeing} explored the fusion of multiple sensor modalities for perception through fog and smoke.
% However, these approaches primarily focus on detection and localization rather than 3D reconstruction and rendering.

Our work bridges these research areas by explicitly modeling temporally varying smoke separately from scene geometry within the 3DGS framework. 
Unlike previous approaches, ours leverages the complementary strengths of RGB and thermal imaging to achieve 3D reconstruction and smoke removal without requiring paired training data.

\section{Method}
\label{sec:method}

\begin{figure}[htbp]
    \centering
    \includegraphics[width=0.8\linewidth]{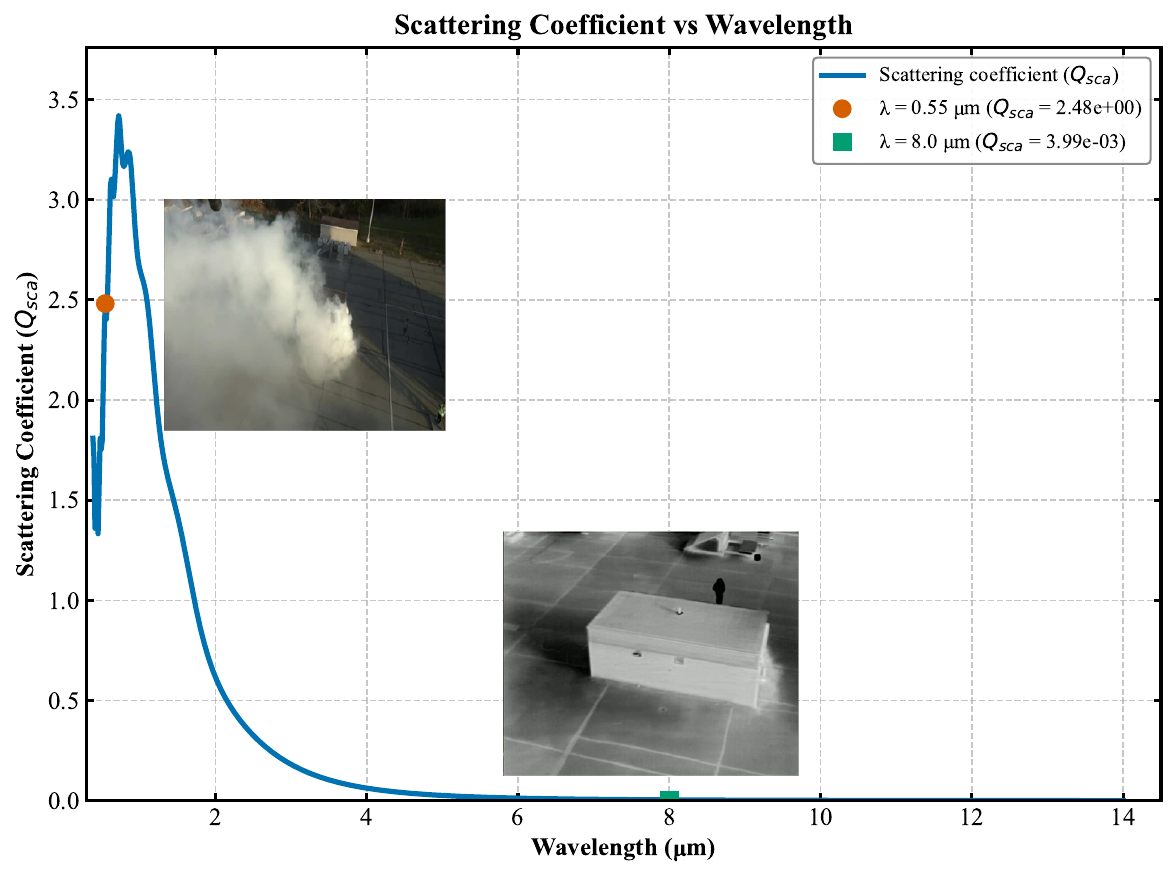}
    \caption{Scattering coefficient as a function of wavelength (in $\mu$m). We calculate the scattering coefficient using size, refractive index using standard values for organic matter found in smoke particles \cite{alves2012influence}. 
    The scattering coefficient is significantly higher in the visible spectrum (0.38--0.7 $\mu$m) compared to the long-wave infrared spectrum (8--14 $\mu$m).
    Inset images which are taken from a drone at roughly the same time illustrate this effect: in visible light (left), smoke strongly obscures the scene, while in thermal infrared (right), the underlying structure is clearly visible.
    }
    \label{fig:scattering}
\end{figure}

\begin{figure*}[!t]
    \centering
    \includegraphics[width=\textwidth]{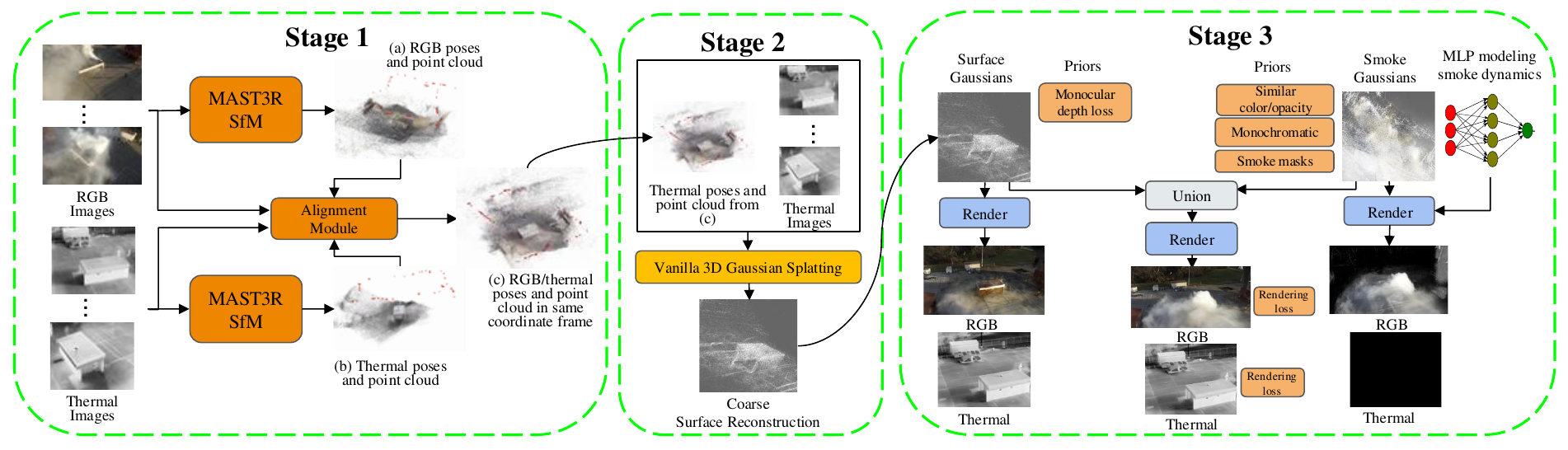}
    \caption{An overview of our method, SmokeSeer. The framework consists of three primary stages: (1) Camera pose estimation and smoke segmentation, (2) Initial surface reconstruction from thermal images, and (3) Joint optimization of surface and smoke plume using both RGB and thermal images.}
    \label{fig:overview}
\end{figure*}

We introduce \emph{SmokeSeer}, a framework for simultaneous 3D scene reconstruction and smoke removal using RGB-thermal image pairs. 
Our approach leverages the complementary strengths of RGB (texture-rich) and thermal (smoke-penetrating) modalities to address the challenges of dense, dynamic smoke in safety-critical applications. Our method comprises three stages (Figure~\ref{fig:overview}): (1) camera pose estimation and smoke segmentation, (2) initial surface reconstruction from thermal images, and (3) joint optimization of surface and smoke using both RGB and thermal images. 

\subsection{Use of thermal images}
\label{subsec:thermal}
Mie theory \cite{mie_theory} provides a framework for understanding how 
different types of particles---such as smoke particles---interact with electromagnetic radiation at different wavelengths. 
For smoke particles of a given size and refractive index, we can use the Mie theory equations to characterize their wavelength-dependent scattering behavior, 
as illustrated in Figure~\ref{fig:scattering}. This analysis reveals a crucial insight: smoke particles predominantly scatter wavelengths in the visible spectrum (0.38--0.7 $\mu$m), 
where RGB cameras operate. However, in the long-wave infrared (LWIR) spectrum (8--14 $\mu$m) 
utilized by thermal cameras, scattering effects from smoke particles are negligible. 
This property allows thermal imaging to penetrate smoke and reveal underlying surface geometry otherwise obscured in RGB imagery. 

In practice, smoke exhibits two key thermal behaviors. First, smoke is largely transparent in the long-wave infrared (LWIR) spectrum because heat dissipates rapidly as smoke moves away from the fire source. 
This transparency enables thermal cameras to capture clear views of scene geometry even when RGB cameras are completely occluded by dense smoke. 
However, in regions extremely close to the fire source, hot smoke can become emissive and appear as a thermal source rather than remaining transparent, 
which our method accounts for in the joint optimization stage.

\subsection{Background on 3D Gaussian splatting}

\label{subsec:background}

% Recently, 3D Gaussian splatting \cite{kerbl3Dgaussians} (3DGS) has emerged as a powerful scene representation that enables faster training and real-time rendering by 
% leveraging rasterization techniques. 
Given a collection of posed images $\{I_k\}_{k=1}^{K}$, $I_k \in \mathbb{R}^{H \times W}$ captured from a scene, 3DGS aims to reconstruct a representation $\mathcal{G}$ of the scene as a set of 3D Gaussians $\mathcal{G}=\{g_i\}$. 
Each Gaussian primitive $g_i$ is characterized by a center position $\boldsymbol{\mu}_i$, 
a symmetric positive-definite covariance matrix $\boldsymbol{\Omega}_i$, 
an alpha value $\alpha_i$, and appearance attributes encoded using spherical 
harmonic coefficients $\boldsymbol{h}_i$ \cite{ramamoorthi2001efficient}. 
Unlike approaches requiring different representations 
for surfaces (e.g., meshes or implicits) and volumes (e.g., voxel grids), 
Gaussian primitives can represent both surfaces and smoke, 
simplifying optimization and rendering.

\subsection{Modeling scattering media using Gaussians}
\label{subsec:model}

We decompose the smoke-filled scene into two sets of Gaussians: surface Gaussians $\mathcal{G}$ 
representing surfaces in the scene, and smoke Gaussians $\mathcal{S}$ capturing the dynamic smoke plume. Before detailing these sets, we explain how to render images using Gaussian primitives.

We first define the transmittance function, which is central to volumetric rendering. For a ray $\mathbf{r}(t) = \mathbf{o} + t\mathbf{d}$ starting at position $\mathbf{o}$ in direction $\mathbf{d}$, the transmittance $T_{\sigma}(t)$ represents the probability that the ray travels from its origin to point $\mathbf{r}(t)$ without obstruction. It is defined as:
\begin{equation}
T_{\sigma}(t) \coloneqq \exp\left(-\int_{t_n}^{t} \sigma(s)  ds\right),
\label{eq:transmittance}
\end{equation}
\noindent where $\sigma(s)$ is the density function along the ray, and $t_n$ is the near-plane distance. In a scene with both surfaces and smoke, we have two density functions: $\sigma(t)$ for surfaces and $\sigma_\mathrm{s}(t)$ for smoke. The combined transmittance is:
\begin{align}
T_{\sigma + \sigma_\mathrm{s}}(t) &= \exp\left(-\int_{t_n}^{t} [\sigma(s) + \sigma_\mathrm{s}(s)] \, ds\right) \\ &= T_{\sigma}(t) \cdot T_{\sigma_\mathrm{s}}(t),
\label{eq:combined_transmittance}
\end{align}
\noindent which represents the probability of the ray reaching point $\mathbf{r}(t)$ without hitting either a surface or smoke particles.

Chen et al. \cite{Chen2024dehazenerf} have shown that the volume rendering equation for a scene with mixed density takes the form:
\begin{equation}
    \begin{aligned}
    C(\mathbf{r}, \mathbf{d}) &=\ 
    \underbrace{\int_{t_n}^{t_0} c(\mathbf{r}(t), \mathbf{d}) \sigma(t) T_{\sigma + \sigma_\mathrm{s}}(t) \, dt}_{C_{\text{surface}}} \\
    &+ \underbrace{\int_{t_n}^{t_0} c_\mathrm{s}(\mathbf{r}(t)) \sigma_\mathrm{s}(t) T_{\sigma + \sigma_\mathrm{s}}(t) \, dt}_{C_{\text{smoke}}},
    \end{aligned}
    \label{eq:4}
\end{equation}    
where $t_0$ is the far-plane distance. Our dual Gaussian representation directly maps to this equation, where the surface Gaussians $\mathcal{G}$ correspond to $C_{\text{surface}}$ 
with color $c$ and opacity $\sigma$, whereas the smoke Gaussians $\mathcal{S}$ correspond to $C_{\text{smoke}}$ with color $c_\mathrm{s}$ and opacity $\sigma_\mathrm{s}$. 
Rendering the union $\mathcal{G} \cup \mathcal{S}$ is equivalent to computing the rendering equation \eqref{eq:4}. 

The rendering equation for the clear-view surfaces without smoke interference is given by:
\begin{equation}
C_{\text{clear}}(\mathbf{r}, \mathbf{d}) = \int_{t_n}^{t_0} c(\mathbf{r}(t), \mathbf{d}) \sigma(t) T_{\sigma}(t) \, dt.
\label{eq:5}
\end{equation}
Rendering only the surface Gaussians $\mathcal{G}$ is equivalent to computing the rendering equation \eqref{eq:5}.
By modeling surface and smoke separately, we achieve effective smoke removal 
through selectively rendering only surface Gaussians.

\subsection{Modality-specific representations}
\label{subsec:notation}

Building on the Gaussian representation described in Section~\ref{subsec:background}, we extend the model to handle both 
RGB and thermal modalities. We use $\{I^{\text{RGB}}_k\}_{k=1}^{K_{\text{RGB}}}$ and $\{I^{\text{T}}_k\}_{k=1}^{K_{\text{T}}}$ 
to denote our RGB and thermal image collections respectively, with associated camera poses $P^{\text{RGB}}_k$ and $P^{\text{T}}_k$.

For our dual Gaussian representation:
\begin{itemize}[nosep,leftmargin=*]
    \item Surface Gaussians $\mathcal{G}$ maintain the parameters from Section~\ref{subsec:background} 
    but with modality-specific spherical harmonic coefficients ($\boldsymbol{h}^{\text{RGB}}_i$, $\boldsymbol{h}^{\text{T}}_i$), and modality-shared opacity $\alpha$.

    \item Smoke Gaussians $\mathcal{S}$ have modality-specific spherical harmonic coefficients and opacities ($\alpha^{\text{RGB}}_i \gg \alpha^{\text{T}}_i$), reflecting the physical properties described in Section~\ref{subsec:thermal}. In addition, they are time-varying to capture smoke dynamics. 
\end{itemize}

\subsection{Stage 1: Generating segmentation masks and obtaining poses}
\label{subsec:stage1} 

In this stage, our objective is to estimate camera poses for RGB and thermal images in a common coordinate system. Accurate cross-modal poses are a prerequisite for any multi-view fusion; without them, correspondence and consistency losses are ill-defined.
This task is challenging due to the different sensor responses between these modalities, 
which complicates cross-modal feature matching. Additionally, the featureless appearance and dynamic smoke
in RGB images impede reliable feature extraction.

We address these challenges with a three-step approach:
\begin{enumerate}
\item {\it Smoke segmentation:} We use GroundedSAM~\cite{ren2024grounded}, based on 
SAMv2~\cite{ravi2024sam2segmentimages}, to identify and mask out smoke-affected regions in RGB images. 
Doing so ensures we match only features from reliable, smoke-free areas.

\item {\it Independent 3D reconstructions:} We run MAST3R-SfM~\cite{Duisterhof2024MASt3RSfMAF} independently 
on RGB and thermal images. Using the masks from the previous step, we discard matches in the smoke regions of RGB images. 
Though MAST3R-SfM handles RGB-RGB and thermal-thermal matching well, it struggles with RGB-thermal matching.

\item {\it Cross-modal registration:} We use MINIMA~\cite{jiang2025minima}, which is specialized for cross-modality matching, to establish 2D correspondences between RGB-thermal image pairs. 
We then lift these correspondences to 3D using the 2D-3D mappings from the per-modality calibration. 
Doing so enables the estimation of a similarity transform $T \in \mathrm{Sim}(3)$ that aligns the RGB and thermal coordinate systems.
\end{enumerate}

\subsection{Stage 2: Reconstructing the scene using thermal images}
\label{subsec:stage2} 

In this stage, we obtain a first reconstruction of the scene geometry using only thermal images, which are minimally affected by smoke. We run vanilla 3D Gaussian splatting~\cite{kerbl3Dgaussians} on the thermal sequence, 
which outputs a smoke-free representation of the scene geometry. 
The surface reconstruction is coarse due to the low resolution of thermal images, but serves as a reliable initialization for our surface Gaussians.

\subsection{Stage 3: Fusing RGB-thermal information and refining geometry}
\label{subsec:stage3}

In the final stage, we jointly optimize surface and smoke Gaussian sets using both RGB and thermal images:
\begin{itemize}[nosep,leftmargin=*]
\item {\it Surface Gaussians:} Initialized from Stage 2, these Gaussians remain static and 
maintain identical opacity across modalities. We augment them with spherical harmonic coefficients to capture RGB appearance.
\item {\it Smoke Gaussians:} Randomly initialized within the scene bounds, these Gaussians evolve temporally and 
exhibit modality-dependent opacity, to model smoke's varying visibility in RGB 
versus thermal images (Section~\ref{subsec:thermal}). 
Though in principle we could use Mie theory to model opacities, 
we opt for a more flexible approach with two independent variables for smoke visibility in each modality.
\end{itemize}

\subsubsection{Modeling the dynamic smoke}
\label{subsubsec:dynamic}
Our approach explicitly accounts for the temporal evolution of smoke, which is critical for applications such as firefighting where smoke behavior is dynamic and unpredictable.
Accounting for smoke motion enables more accurate surface reconstruction in areas temporarily occluded by passing smoke, and improves separation of surface and smoke.
We model the dynamics of smoke following the deformable 3D Gaussians framework \cite{yang2023deformable3dgs}.
This framework uses 3D Gaussians in a canonical space, along with a deformation field to model motion over time.
To model this field, we use a multi-layer perceptron (MLP) that takes as input the positions of the 3D Gaussians and a timestep $t$, and outputs offsets in position, scale, and rotation.
These offsets transform the canonical 3D Gaussians to the deformed space at each time. We use a bimodal Gaussian distribution following \cite{lee2024ex4dgs} to model smoke opacity as a function of time.

\subsubsection{Priors on properties of smoke Gaussians}
\label{subsubsec:priors}

To facilitate accurate surface-smoke separation and modeling of realistic smoke behavior, 
during optimization we use priors motivated by physical properties of smoke:

\begin{itemize}
\item {\it Smoke consistency:} We minimize variance in opacity and color across smoke Gaussians:
\begin{equation}
    L_{\mathrm{smoke\_alpha}} = \text{Var}(\{\alpha_i\}_{i \in \mathcal{S}})
\end{equation}
\begin{equation}
    L_{\mathrm{smoke\_color}} = \text{Var}(\{c_i\}_{i \in \mathcal{S}})
\end{equation}
This prior reflects the physical observation that smoke particles in a local region typically 
have similar optical properties. In real smoke, particles of similar size and composition 
would have nearly identical opacity and scattering properties. 
By enforcing consistency across smoke Gaussians, we prevent unrealistic variations from arising during optimization.
Though the loss should ideally apply to Gaussians in local neighborhoods, 
we found that applying it across all Gaussians works well in practice.

\item {\it Monochromaticity:} We enforce consistent color channels across smoke Gaussians:
\begin{equation}
    L_{\mathrm{mono}} = \sum_{i \in \mathcal{S}} \text{Var}(c_i^\mathrm{R}, c_i^\mathrm{G}, c_i^\mathrm{B}).
\end{equation}
This prior reflects the physical property that smoke typically appears as a neutral gray color.
It prevents our model from generating implausible colored smoke.

\item {\it Depth consistency:} We align the surface Gaussians with monocular depth cues:
\begin{equation}
    L_{\mathrm{depth}} = \|d_i - \hat{d}_i\|,
\end{equation}

where $d_i$ denotes predicted depth on a thermal image using a monocular depth estimation model~\cite{depth_anything_v2} 
and $\hat{d}_i$ is the rendered depth from the surface Gaussians using thermal camera parameters. 
This prior leverages the smoke-penetrating property of thermal imaging (Section~\ref{subsec:thermal}). 
Since thermal images are minimally affected by smoke, they provide reliable depth cues for the underlying surface geometry, helping to prevent surface Gaussians from being incorrectly positioned in smoke-occluded regions.

\item {\it Mask alignment:} The alpha values of smoke Gaussians should be consistent with the masks from Stage 1:
\begin{equation}
    L_{\mathrm{mask}} = \|M_{\mathrm{pred}} - M_{\mathrm{GT}}\|_1, 
\end{equation}
where $M_{\mathrm{pred}}$ and $M_{\mathrm{GT}}$ are the pixel-wise accumulated alpha values of the rendered smoke Gaussians and segmentation masks, respectively. 
This prior ensures spatial consistency between our reconstructed smoke volume and the observed smoke regions in input images. 
It helps constrain the optimization to place smoke Gaussians in only regions with smoke present, and prevent them from appearing in smoke-free ones.
\end{itemize}

The total optimization loss is a weighted sum of these physically-motivated priors and a standard rendering loss for RGB and thermal images:
\begin{align}
    L_{\mathrm{total}} &= \lambda_{\mathrm{render}}L_{\mathrm{render}} + \lambda_{\mathrm{smoke\_alpha}} L_{\mathrm{smoke\_alpha}} \nonumber \\ 
    &+ \lambda_{\mathrm{smoke\_color}}L_{\mathrm{smoke\_color}} + \lambda_{\mathrm{mono}}L_{\mathrm{mono}} \nonumber \\ 
    &+ \lambda_{\mathrm{depth}}L_{\mathrm{depth}} + \lambda_{\mathrm{mask}}L_{\mathrm{mask}}.\label{eq:loss}
\end{align}

This formulation enables separation of scene geometry from smoke, while maintaining 
physical consistency across the RGB and thermal modalities.
\section{Experimental evaluation}
\label{sec:experimental}

\begin{figure*}[!t]
  \centering
  
  % First row - Caterpillar with headers
  \begin{tabular}{@{}c@{}c@{}c@{}c@{}c@{}c@{}}
    Thermal Image & RGB Image & Ground Truth & Ours (Full) & ImgDehaze & Ours (RGB only) \\
    % Third row - House
    \includegraphics[width=0.165\textwidth]{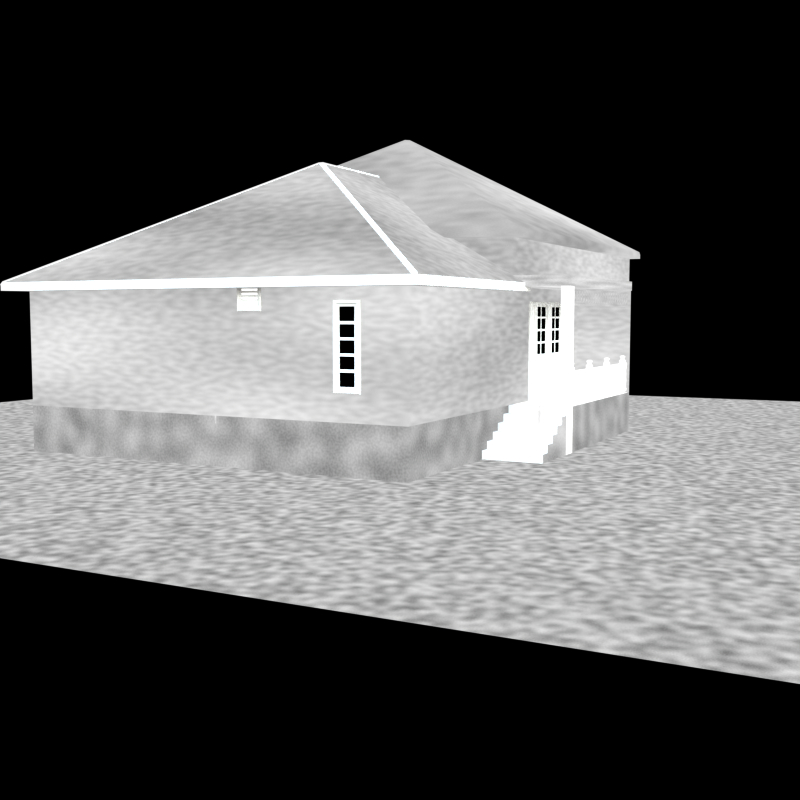}&%
    \includegraphics[width=0.165\textwidth]{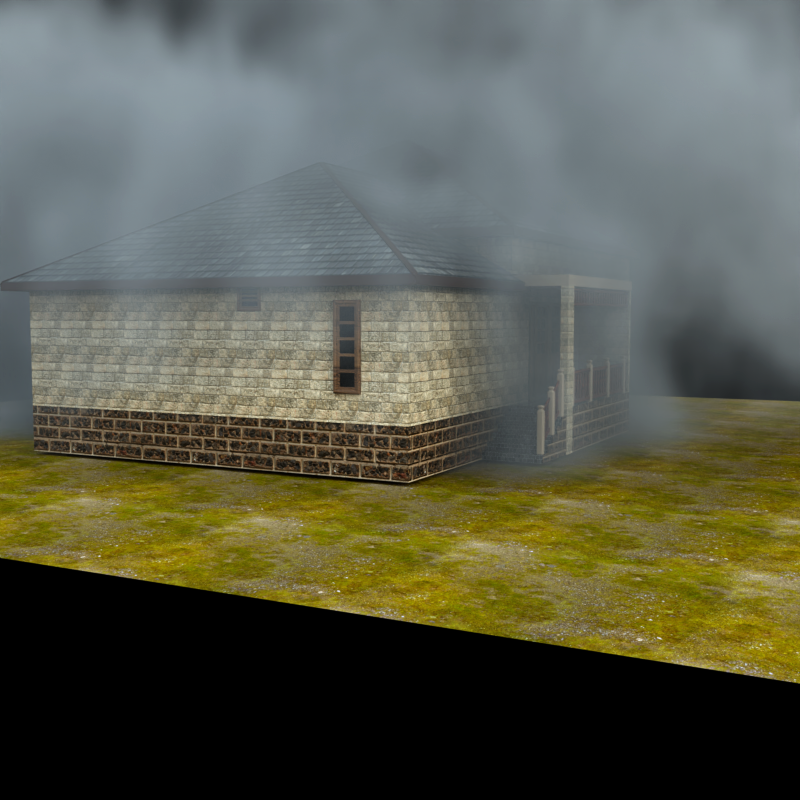}&%
    \includegraphics[width=0.165\textwidth]{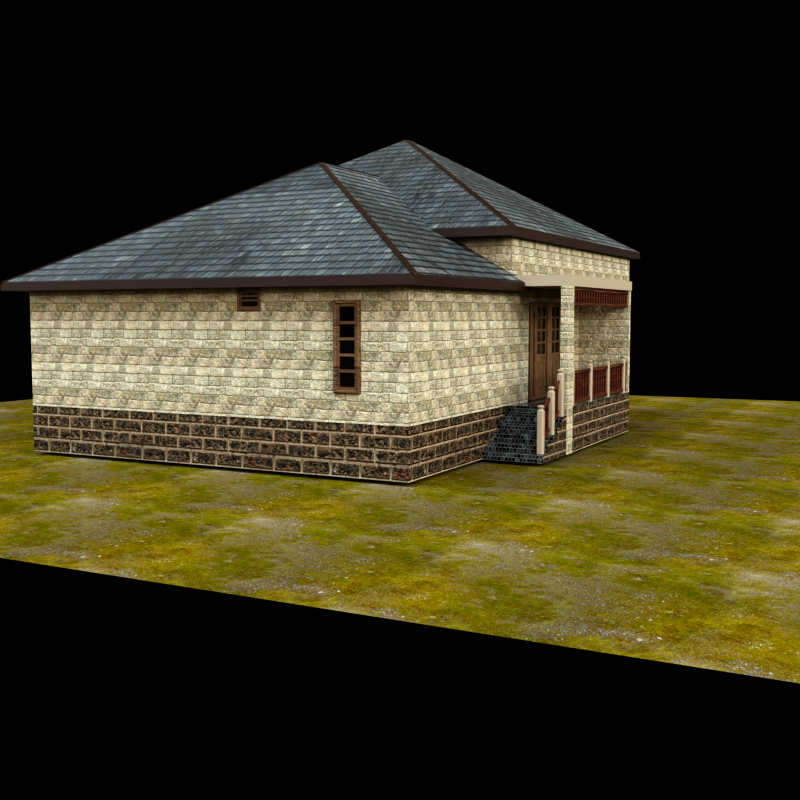}&%
    \includegraphics[width=0.165\textwidth]{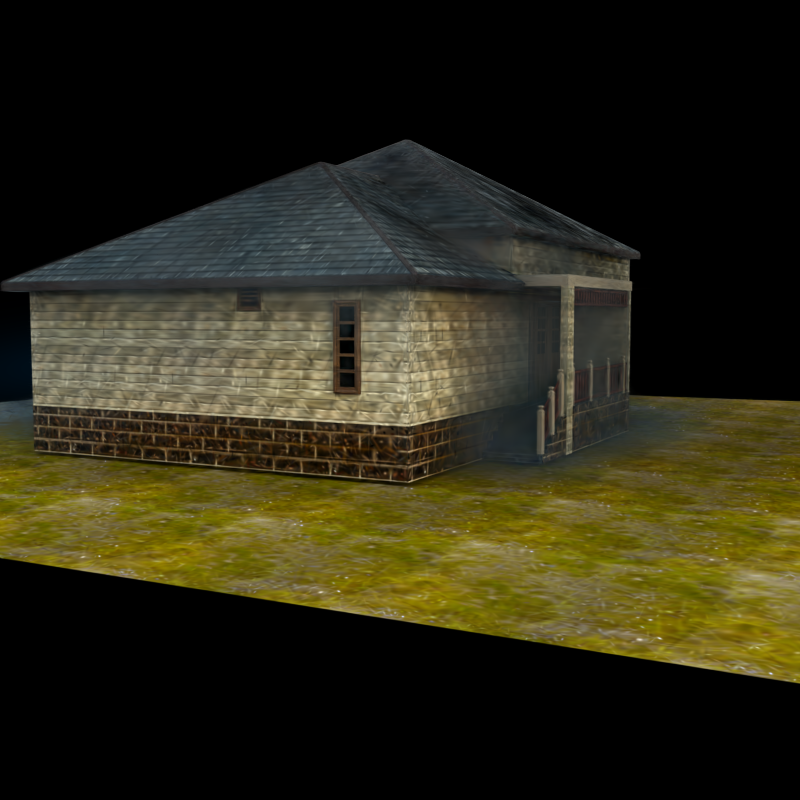}&%
    \includegraphics[width=0.165\textwidth]{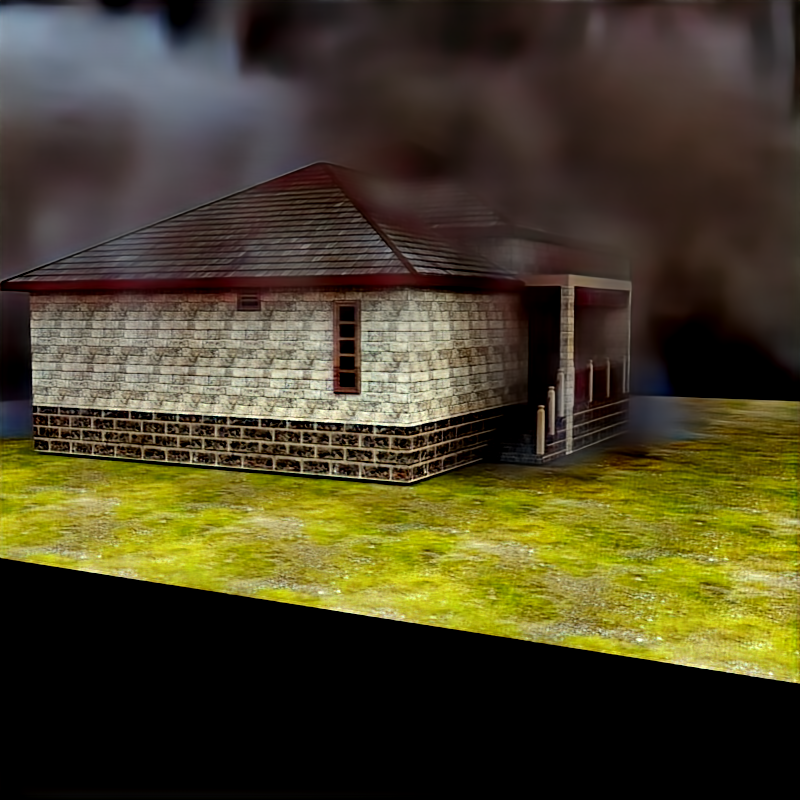}&%
    \includegraphics[width=0.165\textwidth]{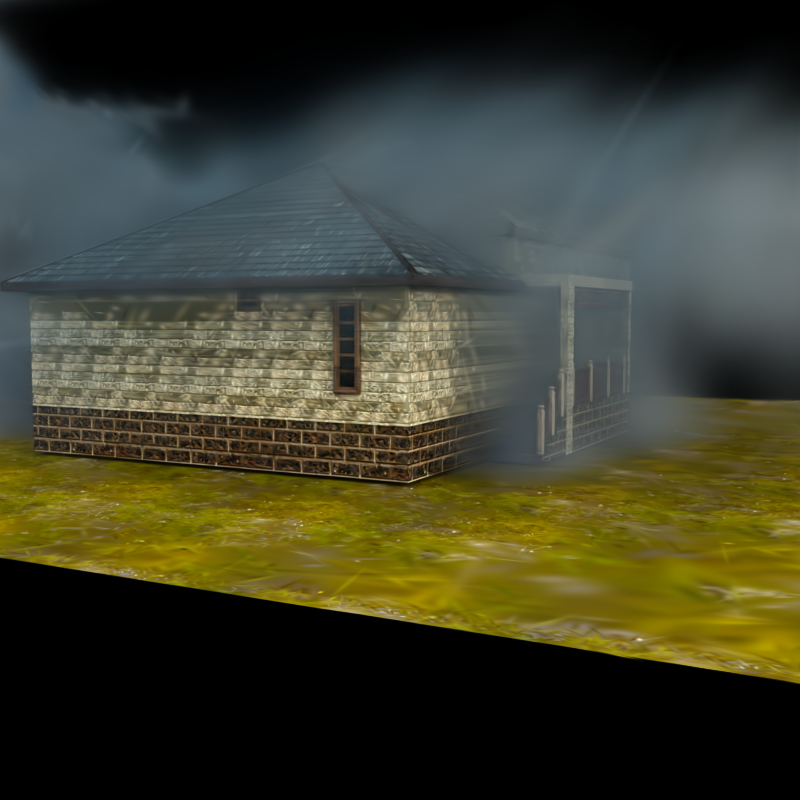}\\[-0.5ex]
    % Fourth row - Mars
    \includegraphics[width=0.165\textwidth]{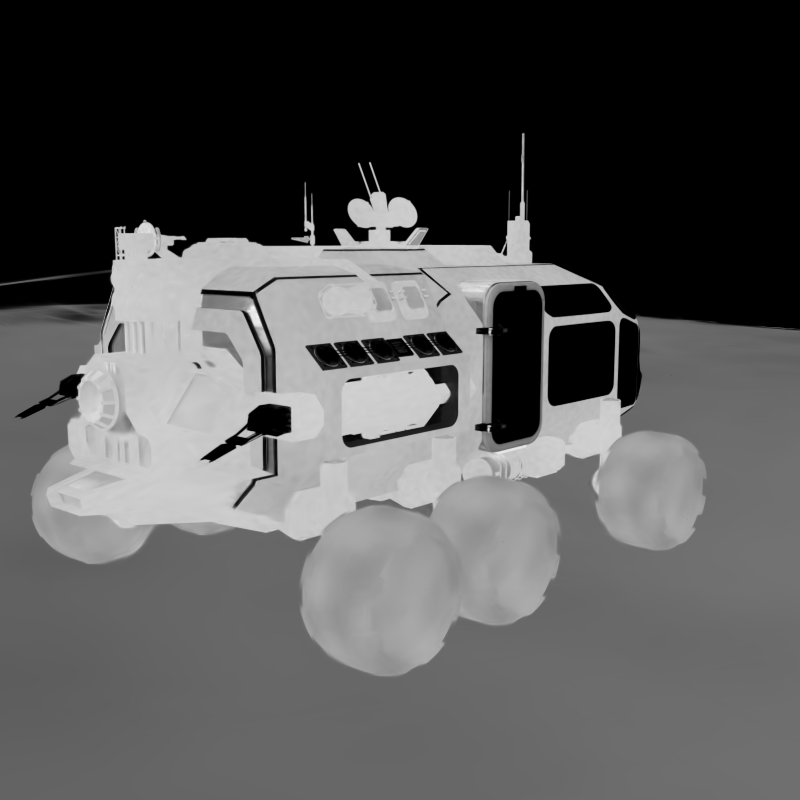}&%
    \includegraphics[width=0.165\textwidth]{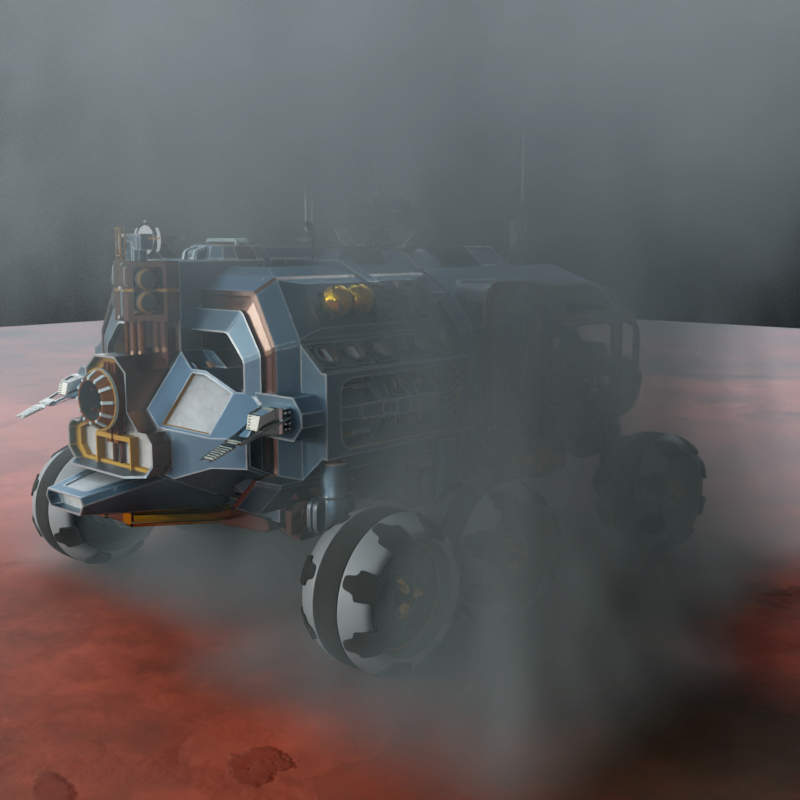}&%
    \includegraphics[width=0.165\textwidth]{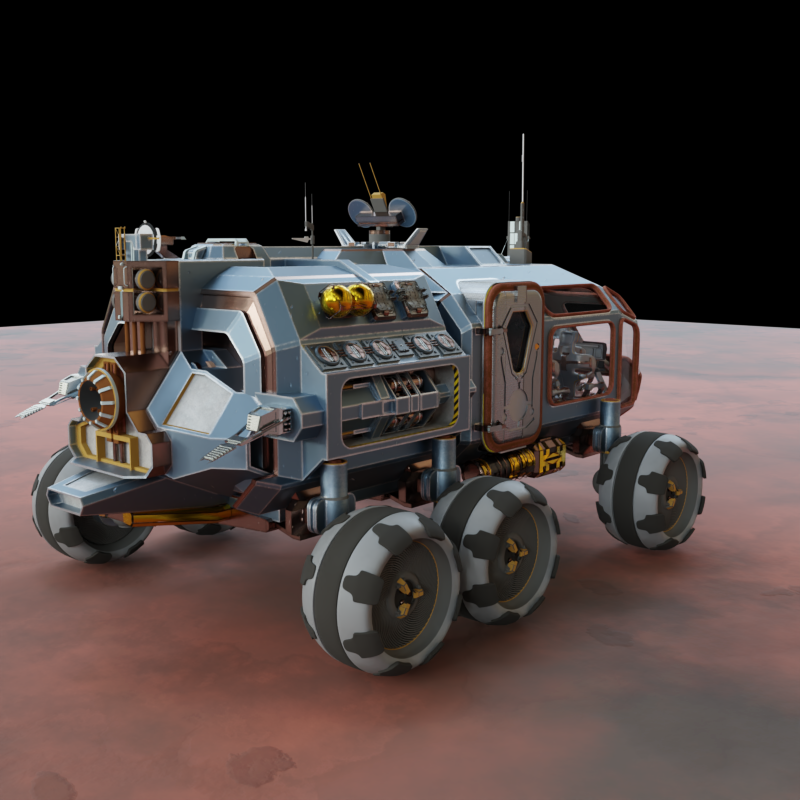}&%
    \includegraphics[width=0.165\textwidth]{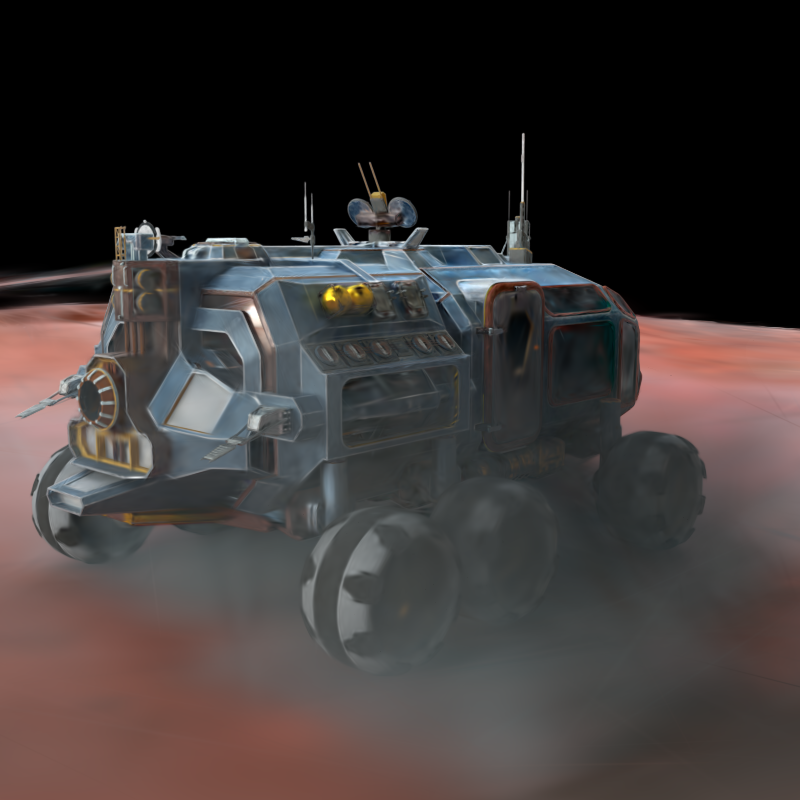}&%
    \includegraphics[width=0.165\textwidth]{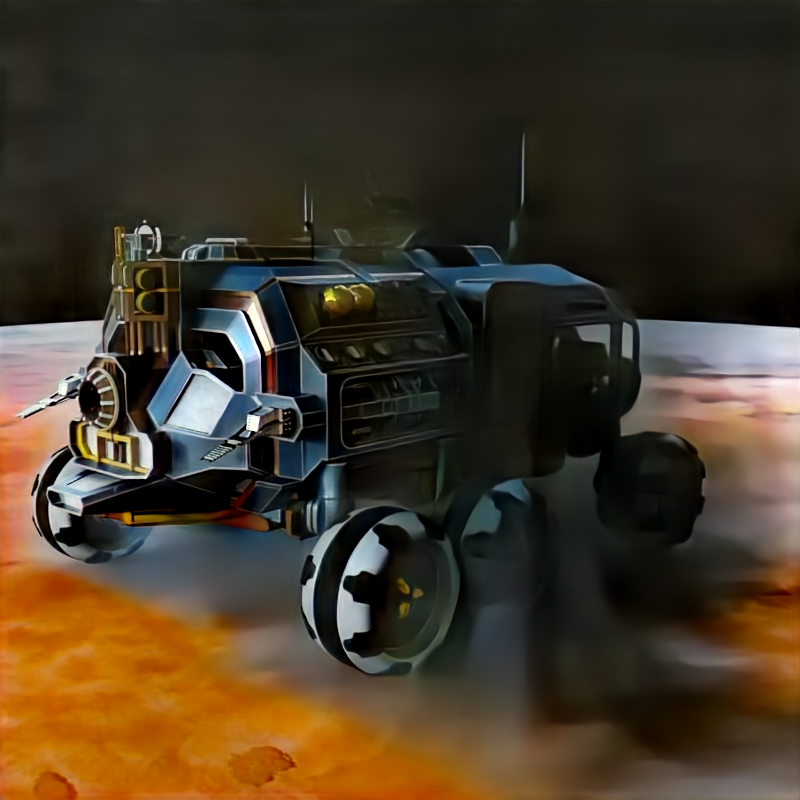}&%
    \includegraphics[width=0.165\textwidth]{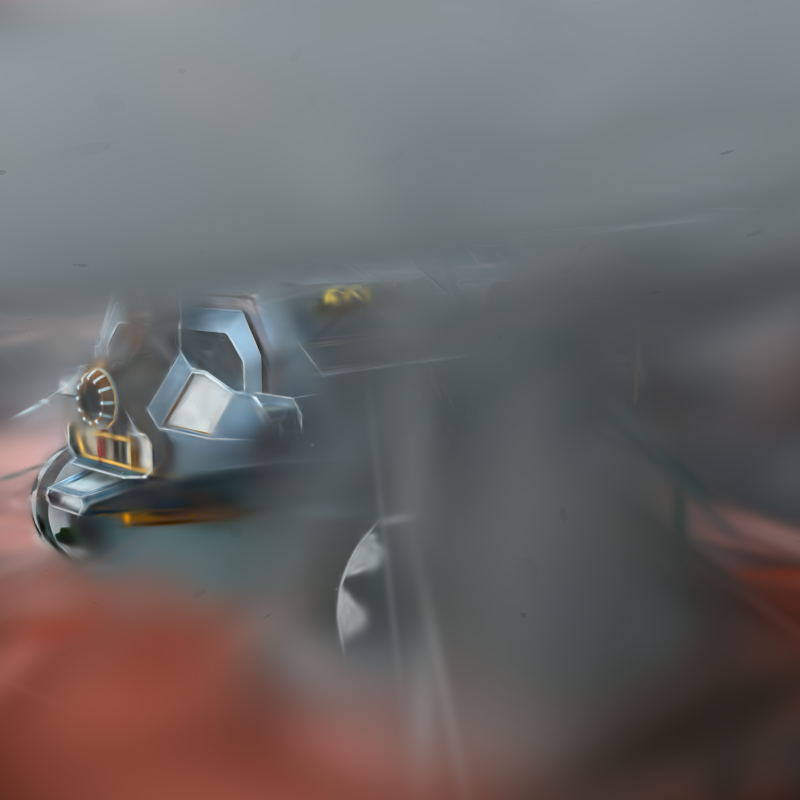} \\ [-0.5ex]

    \includegraphics[width=0.165\textwidth]{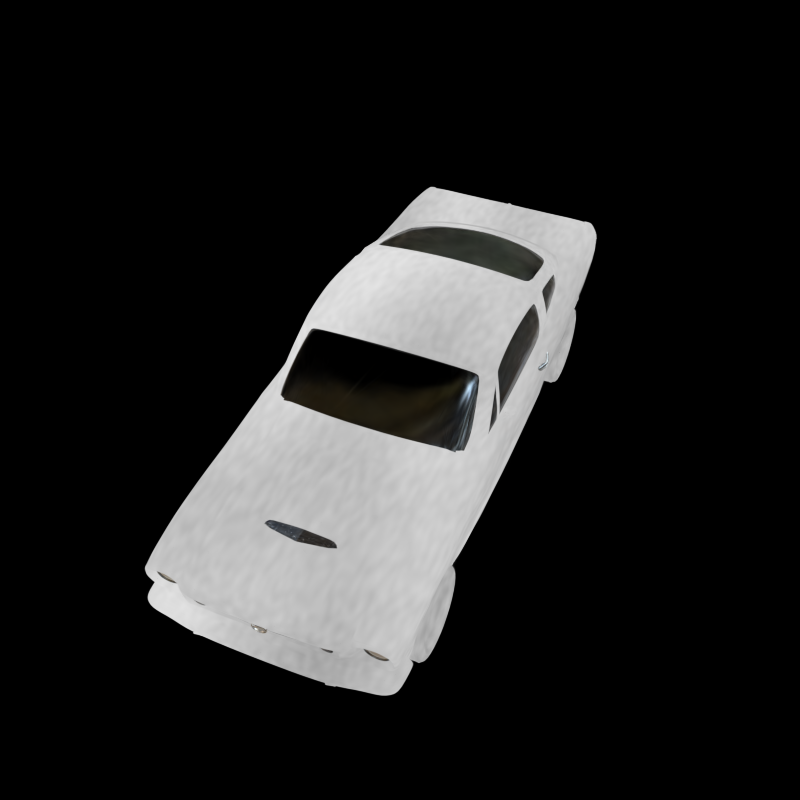}&%
    \includegraphics[width=0.165\textwidth]{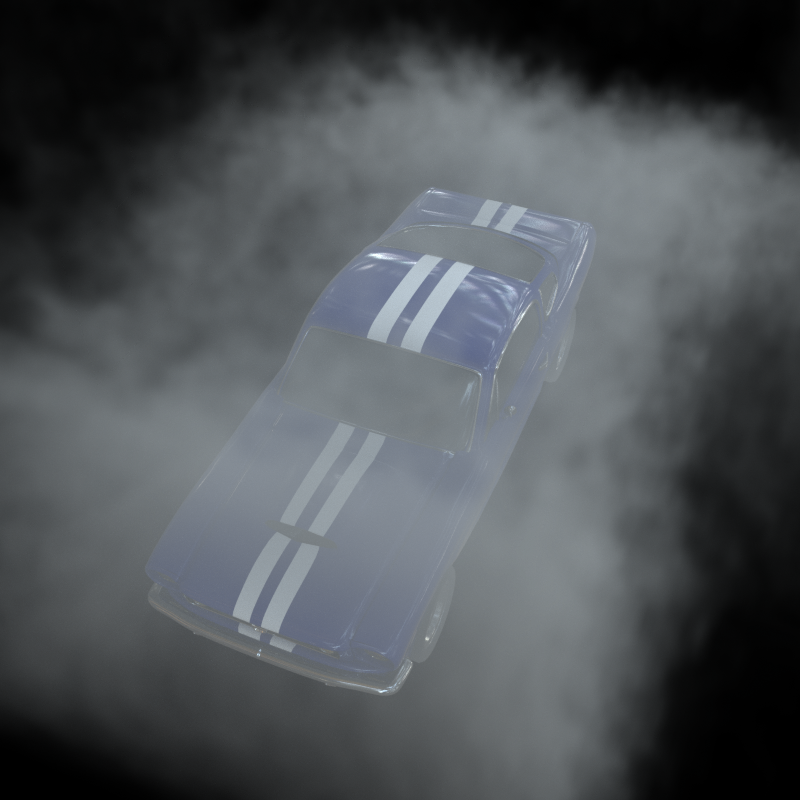}&%
    \includegraphics[width=0.165\textwidth]{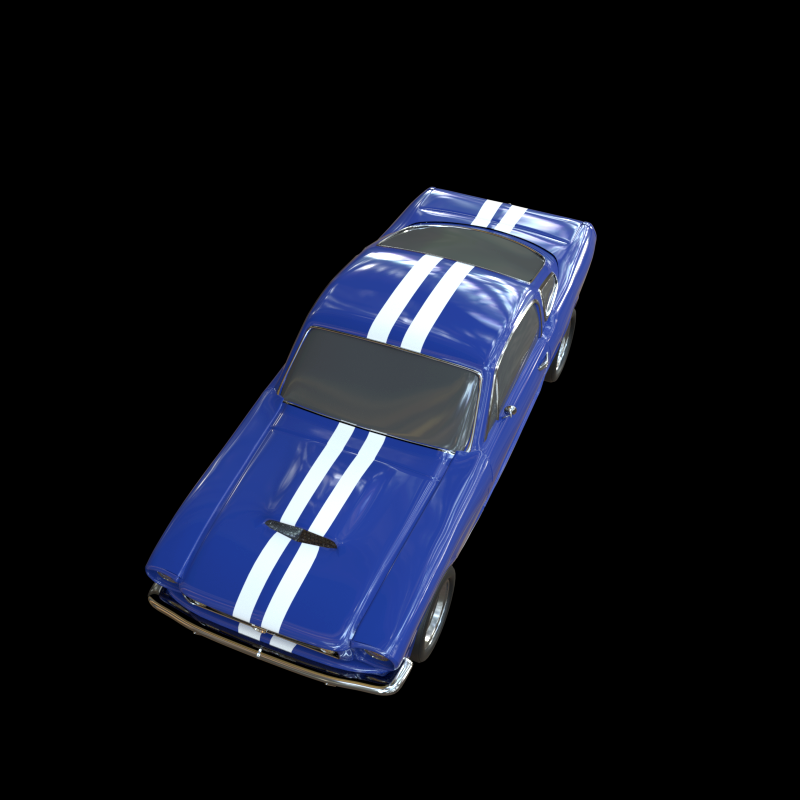}&%
    \includegraphics[width=0.165\textwidth]{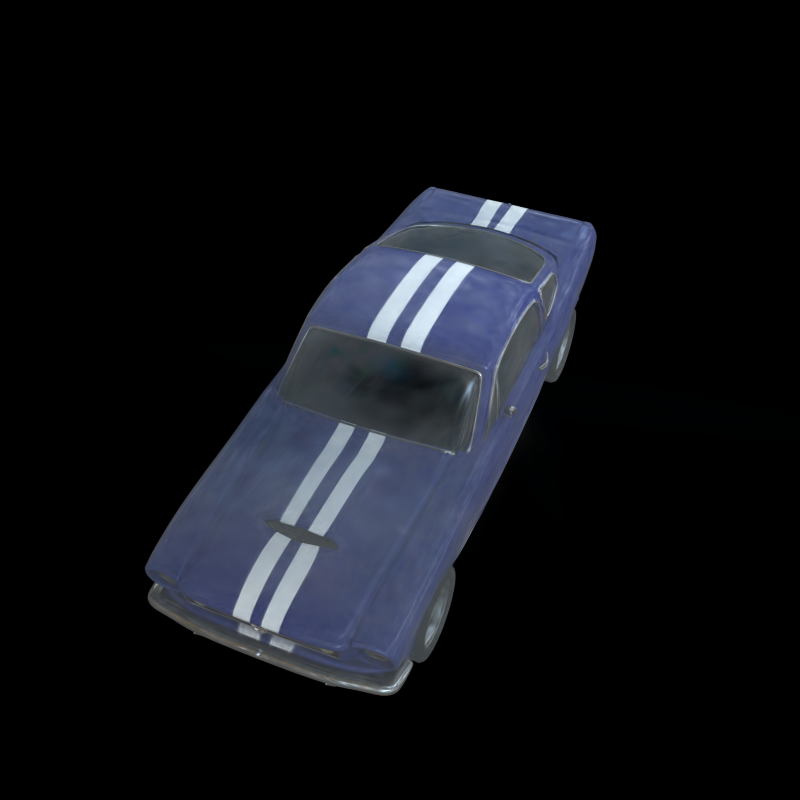}&%
    \includegraphics[width=0.165\textwidth]{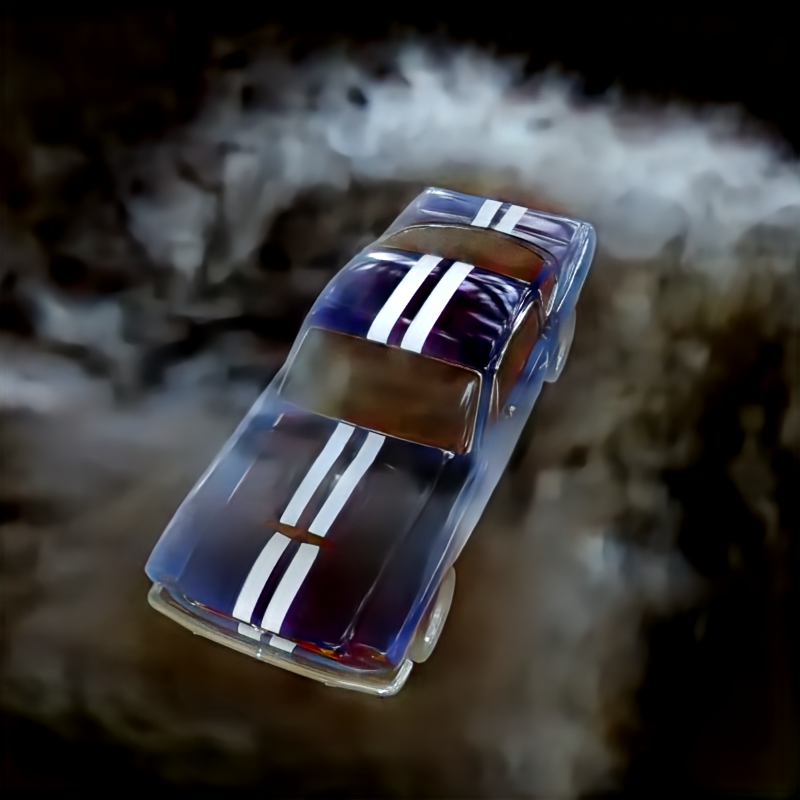}&%
    \includegraphics[width=0.165\textwidth]{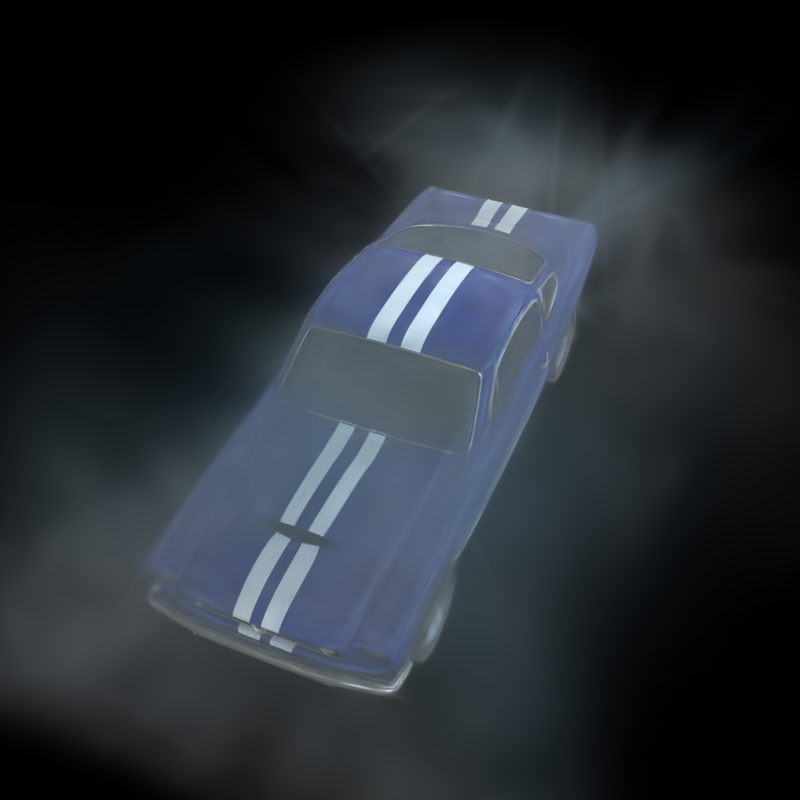}\\[-0.5ex]
    % Fourth row - Mars
    \includegraphics[width=0.165\textwidth]{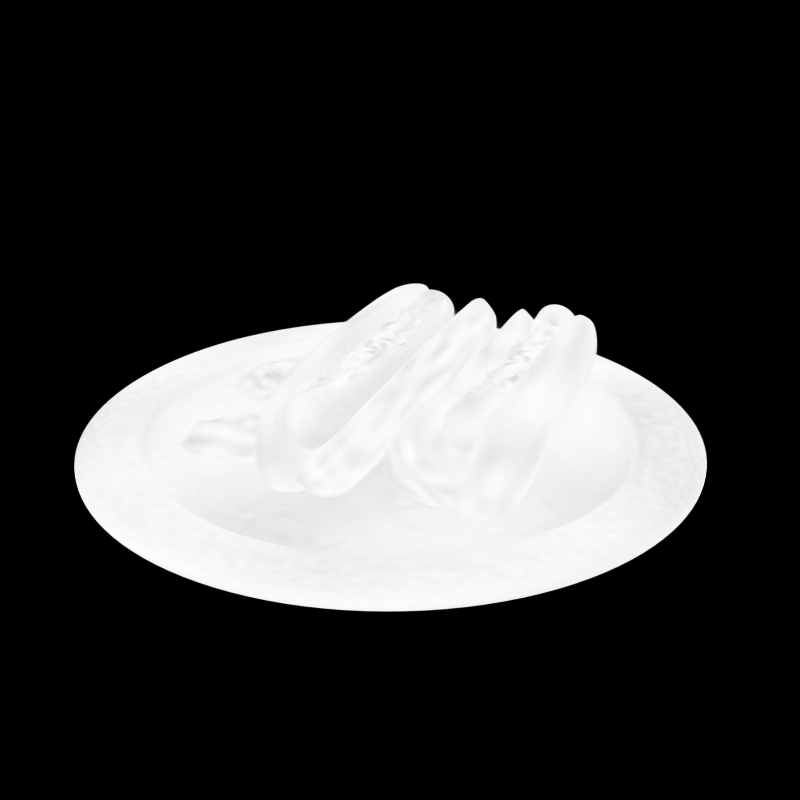}&%
    \includegraphics[width=0.165\textwidth]{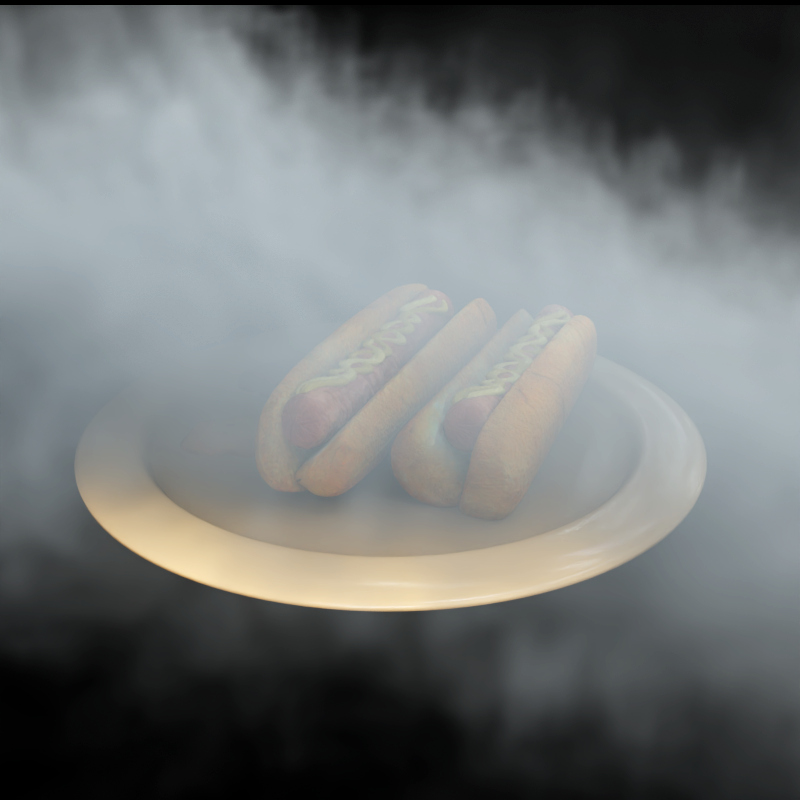}&%
    \includegraphics[width=0.165\textwidth]{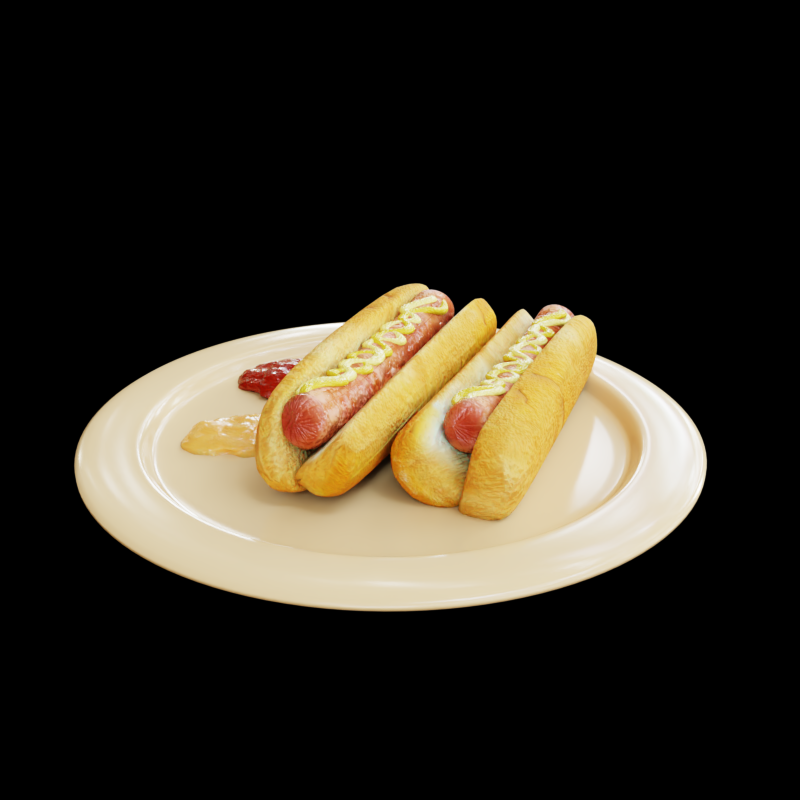}&%
    \includegraphics[width=0.165\textwidth]{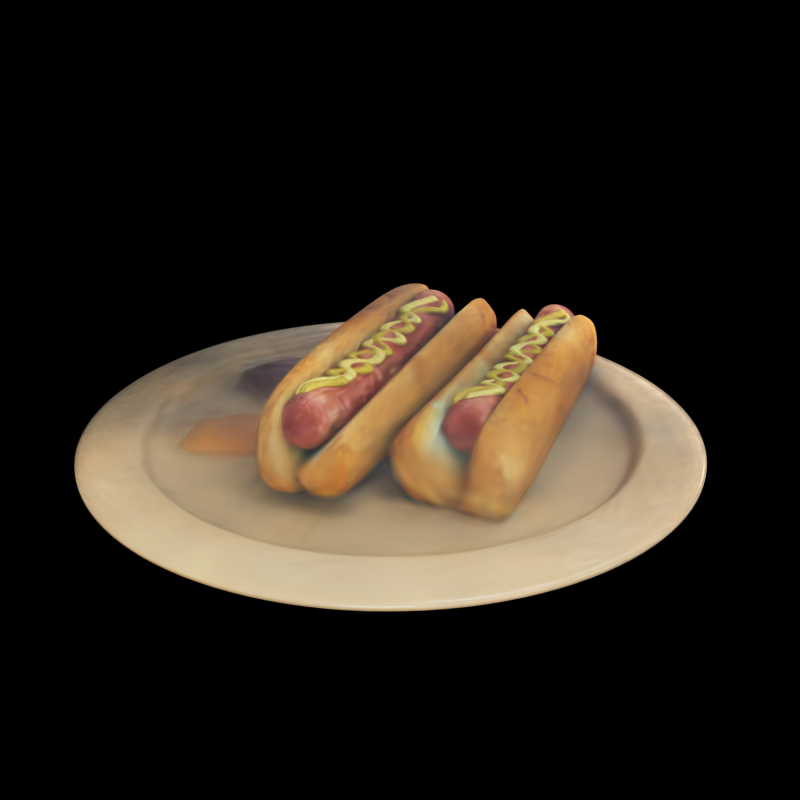}&%
    \includegraphics[width=0.165\textwidth]{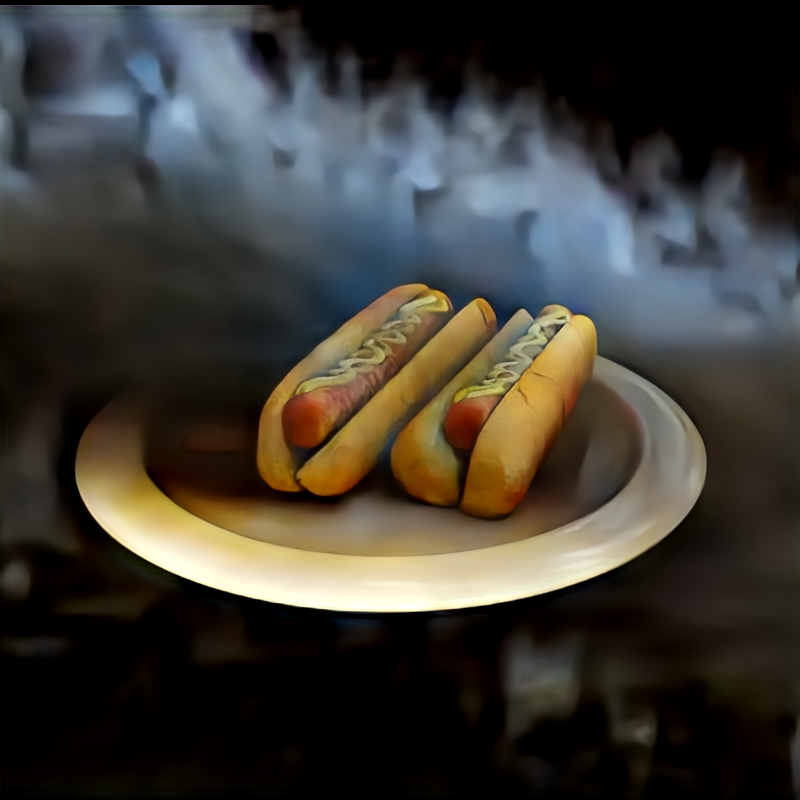}&%
    \includegraphics[width=0.165\textwidth]{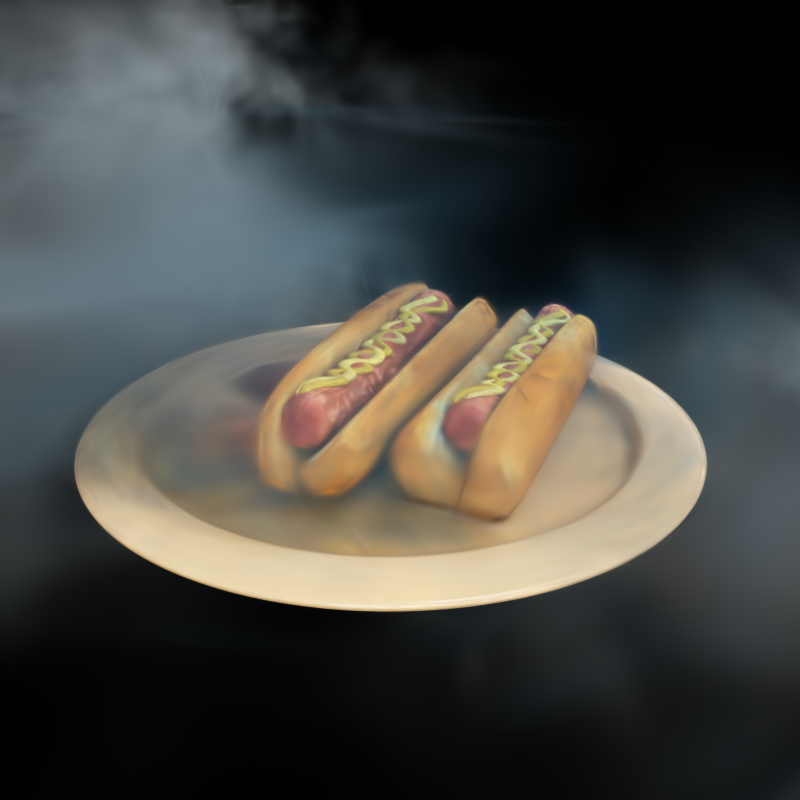}

  \end{tabular}
  
  \caption{Qualitative results on the synthetic dataset. Our full method effectively removes smoke while preserving structural and texture details, outperforming RGB-only approaches.}
  \label{fig:synthetic_qualitative}
\end{figure*}

\begin{figure*}[!t]
  \centering
    \begin{tabular}{@{}c@{}c@{}c@{}c@{}c@{}c@{}}
    Thermal Image & RGB Image & Reference & Ours (Full) & ImgDehaze & Ours (RGB only) \\
    \includegraphics[height=2.3cm, width=0.11955\textwidth]{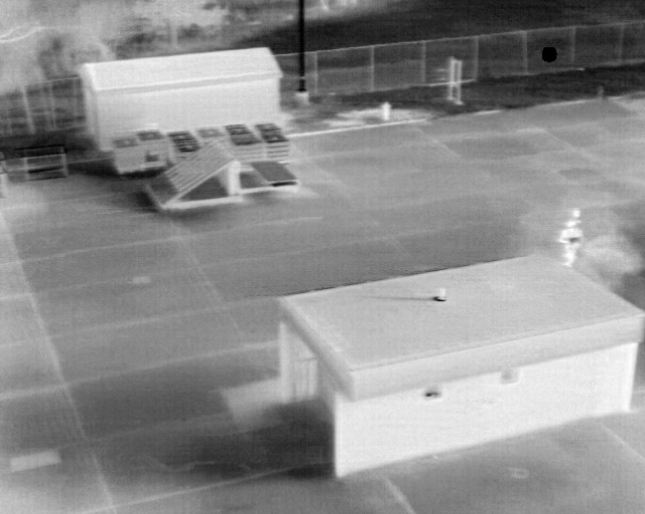}&%
    \includegraphics[height=2.3cm, width=0.1742\textwidth]{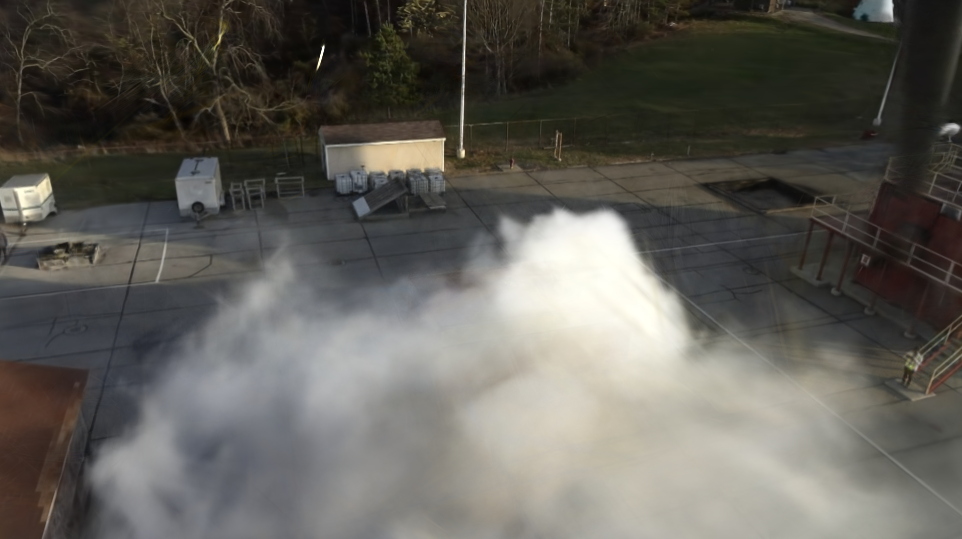}&%
    \includegraphics[height=2.3cm, width=0.1742\textwidth]{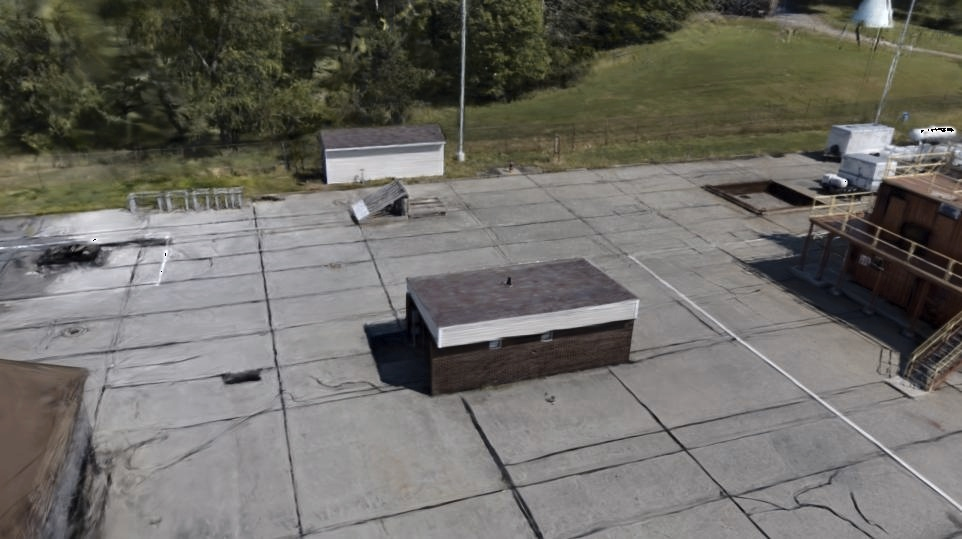}&%
    \includegraphics[height=2.3cm, width=0.1742\textwidth]{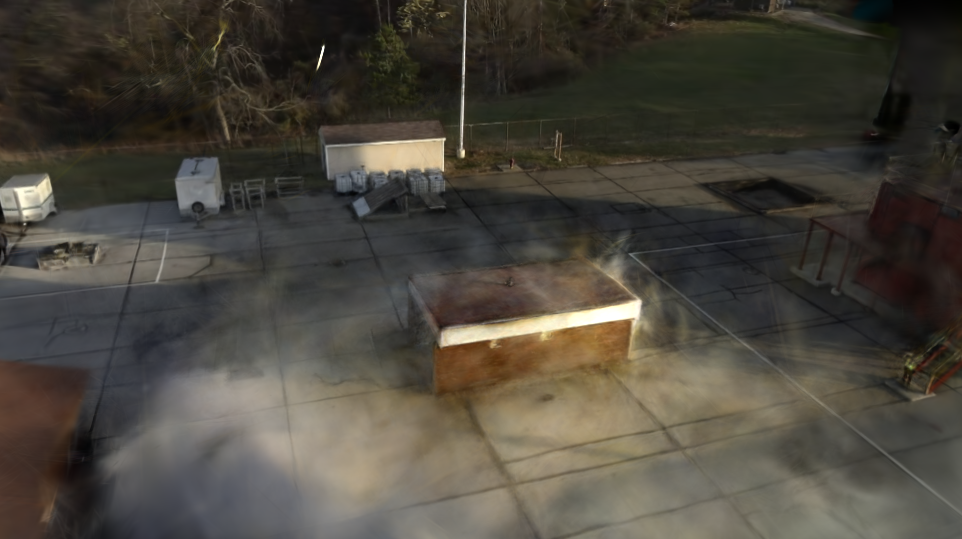}&%
    \includegraphics[height=2.3cm, width=0.1742\textwidth]{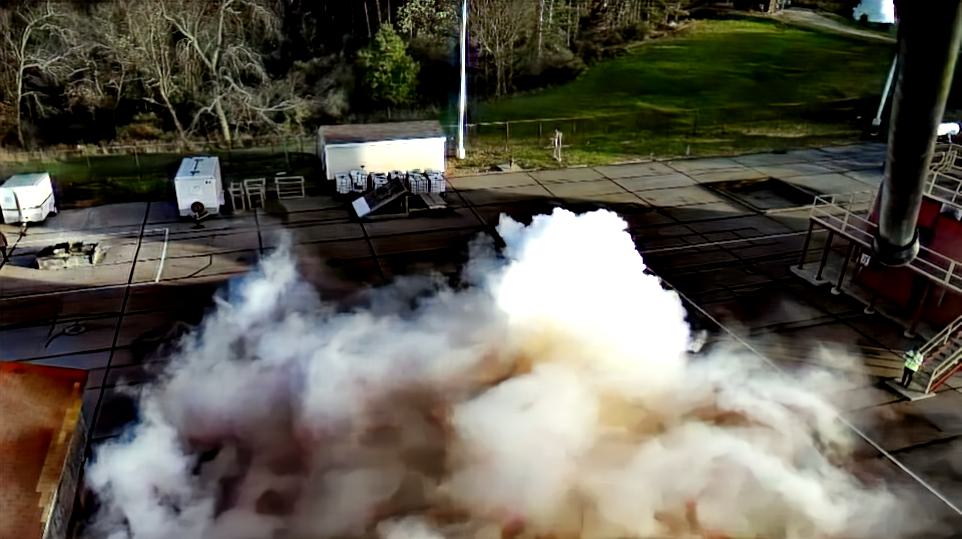}&%
    \includegraphics[height=2.3cm, width=0.1742\textwidth]{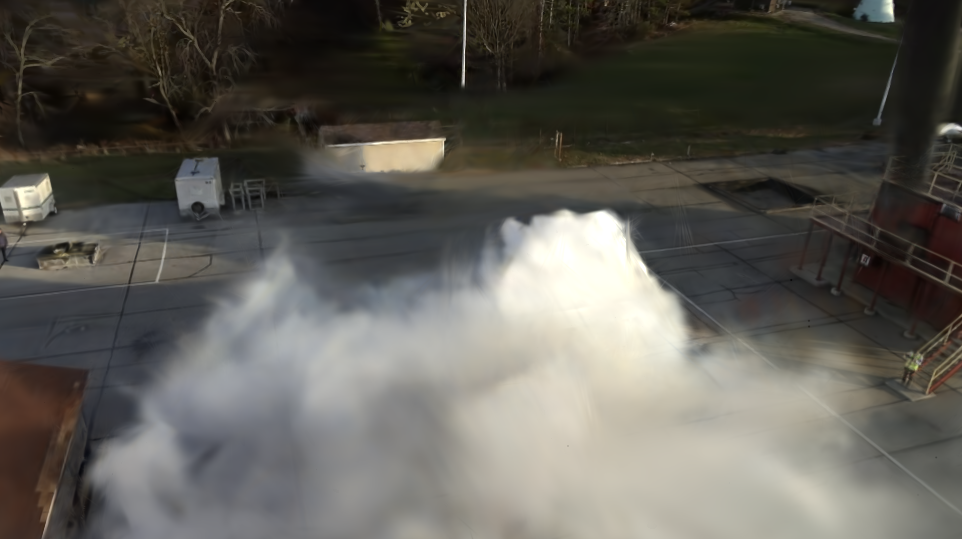}\\[-0.5ex]
    % Second row - Forest
    \includegraphics[height=2.3cm, width=0.11955\textwidth]{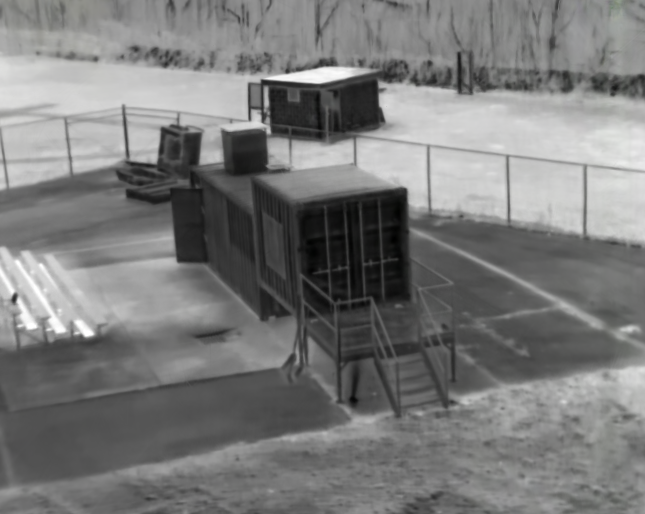}&%
    \includegraphics[height=2.3cm, width=0.1742\textwidth]{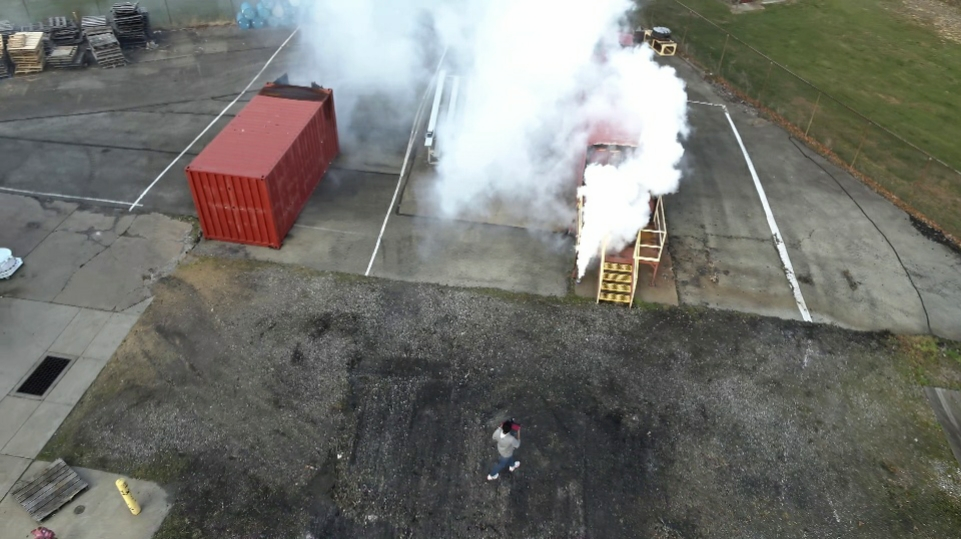}&%
    \includegraphics[height=2.3cm, width=0.1742\textwidth]{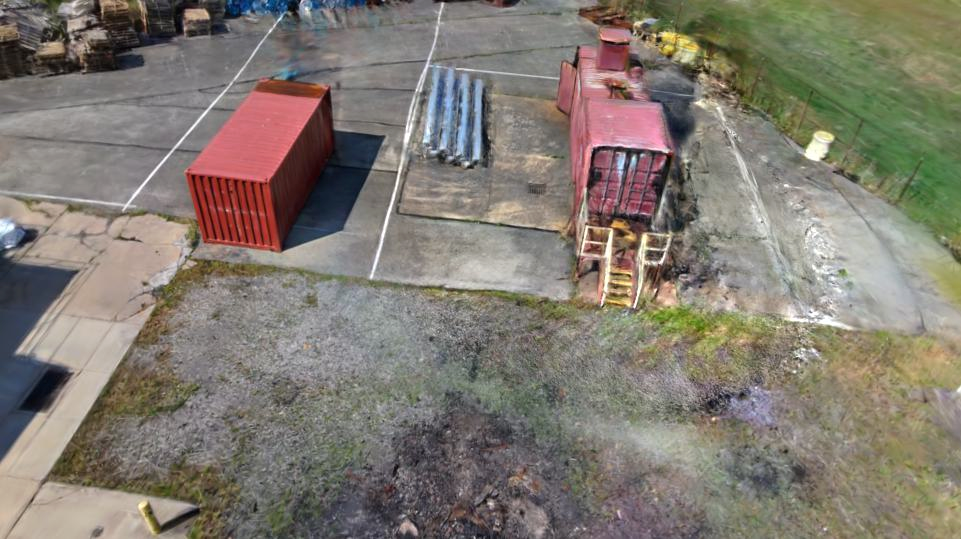}&%
    \includegraphics[height=2.3cm, width=0.1742\textwidth]{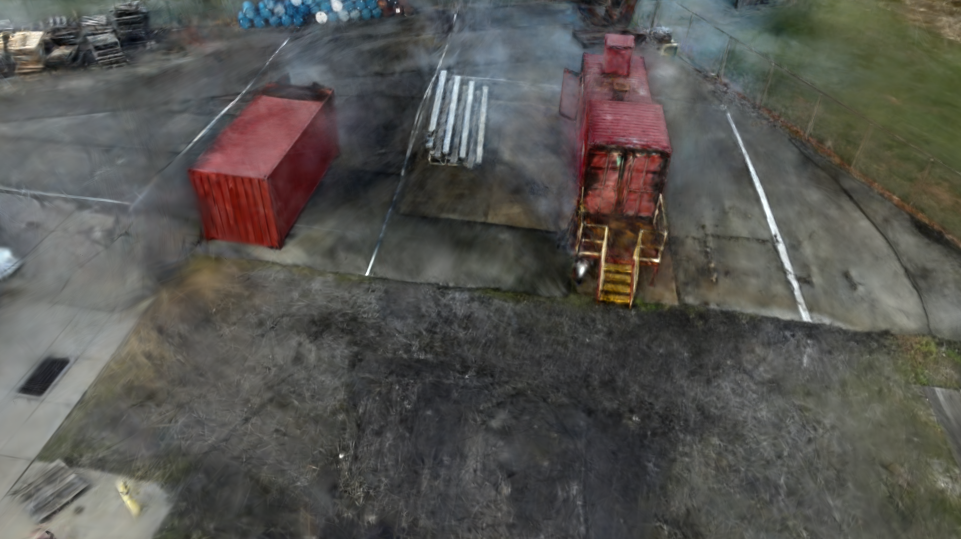}&%
    \includegraphics[height=2.3cm, width=0.1742\textwidth]{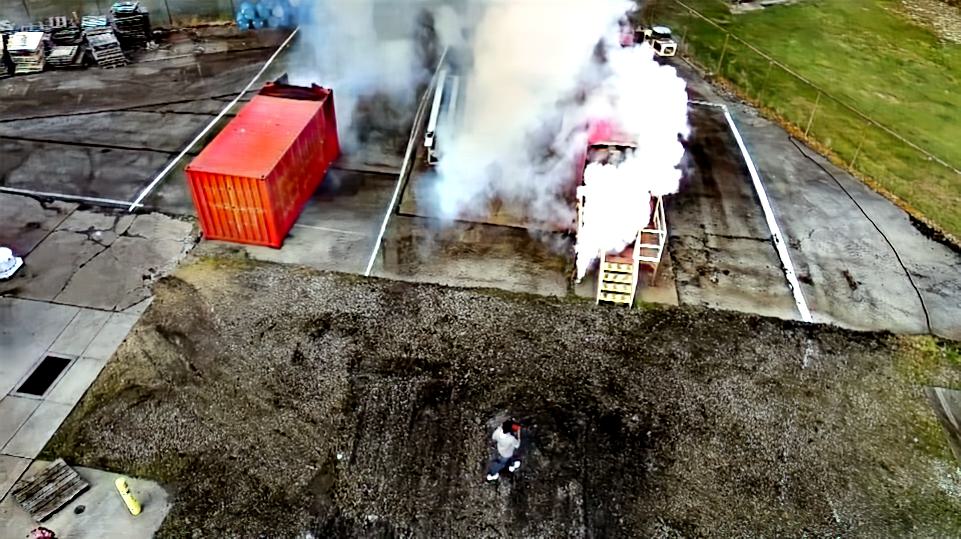}&%
    \includegraphics[height=2.3cm, width=0.1742\textwidth]{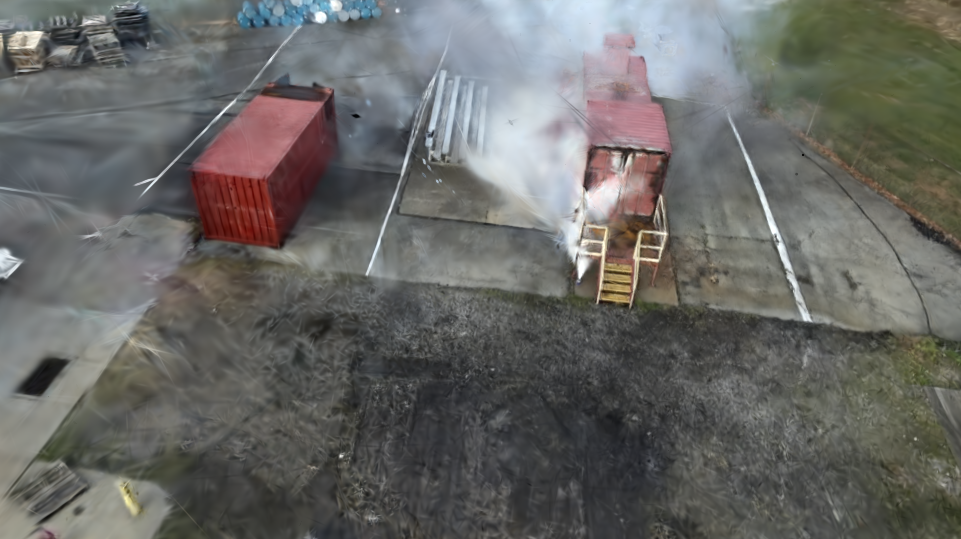}\\[-0.5ex]
   \end{tabular}
   
  \caption{Qualitative results on the real-world dataset. Our method successfully removes smoke in challenging real-world conditions while preserving scene details.}
  \label{fig:real_qualitative}
\end{figure*}

We evaluate our method on synthetic and real-world datasets to demonstrate its effectiveness 
for smoke removal and 3D scene reconstruction. We compare against state-of-the-art methods, and validate our design choices through ablation studies. We provide implementation details in the supplement, and video results on the project website.

\subsection{Datasets}
\label{subsec:datasets}

\paragraph{Synthetic dataset.}
For quantitative evaluation with ground truth, we create a synthetic dataset using Blender's Mantaflow \cite{mantaflow} 
smoke simulator. The dataset comprises 10 scenes: 5 object-level scenes from the NeRF synthetic dataset \cite{mildenhall2020nerf}, 
and 5 large-scale scenes. For each scene, we generate 150 RGB and thermal frames with dynamic smoke.

\paragraph{Real-world dataset.}
In collaboration with our county's fire department, we collected a real-world dataset using a Spirit drone 
equipped with roughly co-located RGB and thermal cameras. There is no time synchronization between the frames captured by the RGB and thermal cameras on the drone, which makes relative pose estimation challenging. 
% The RGB camera captures images at 4K resolution, whereas the thermal camera (FLIR Boson) captures images at a lower resolution of 640$\times$512. Data was collected in two environments using controlled smoke bombs that generate 
% dense smoke representative of firefighting scenarios. 
We do not report quantitative metrics for the real-world dataset as obtaining true ground truth is challenging in such environments. 
Instead, we provide an approximation which we refer to as ``Reference'' in the figures. 
We provide more details in the supplement.  
% To create this reference, we collected additional smoke-free RGB images of the same scenes in a separate 
% drone flight. We then performed the following steps: (1) reconstructed the smoke-free scene using 3DGS, 
% (2) obtained the camera poses of smoke-filled and smoke-free image sets in the same coordinate frame, and 
% (3) rendered novel views of the smoke-free reconstruction using the camera poses from the smoke-filled sequence. 
% These reference images serve as an approximate benchmark, though they are not perfect ground truth since the poses are noisy and environmental conditions change between captures.
This dataset presents several challenges not found in synthetic data, including: imperfect alignment between RGB and thermal cameras, 
unpredictable smoke motion due to wind, and motion blur from drone movement. 
These factors make our real-world dataset a rigorous benchmark for evaluating the 
practical utility of smoke removal algorithms in safety-critical applications.

\begin{figure*}[t]
  \centering
  \begin{tabular}{@{}c@{}c@{}c@{}c@{}c@{}c@{}}
    w/o $\mathcal{L}_{\mathrm{smoke\_alpha}}$ &
    w/o $\mathcal{L}_{\mathrm{smoke\_color}}$ &
    w/o $\mathcal{L}_{\mathrm{mono}}$ &
    w/o $\mathcal{L}_{\mathrm{depth}}$ &
    w/o $\mathcal{L}_{\mathrm{mask}}$ &
    Full method \\
      \includegraphics[width=0.165\linewidth]{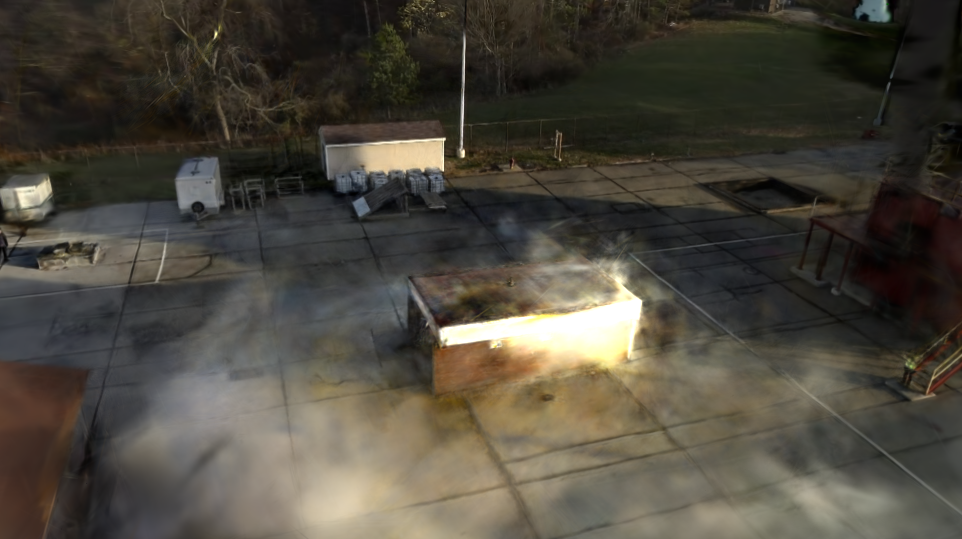} &
      \includegraphics[width=0.165\linewidth]{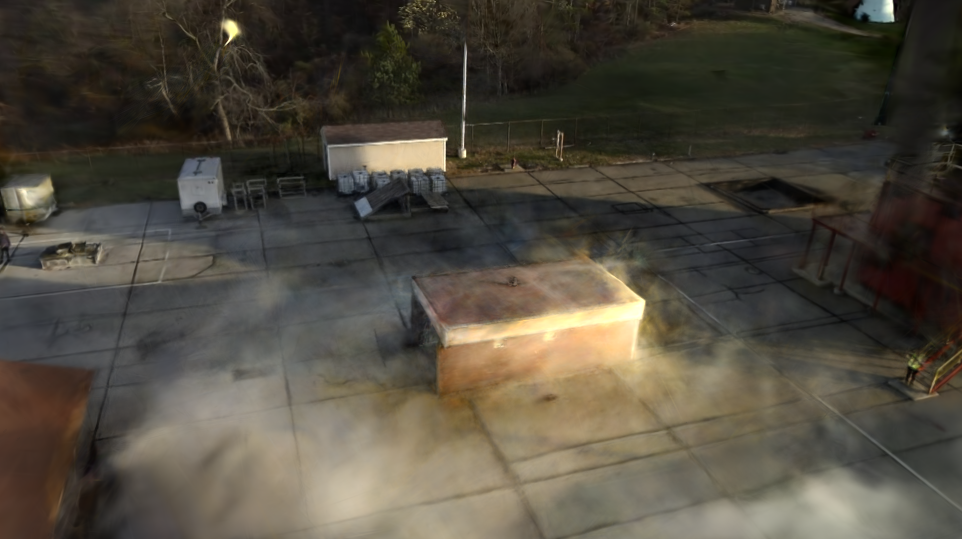} &
      \includegraphics[width=0.165\linewidth]{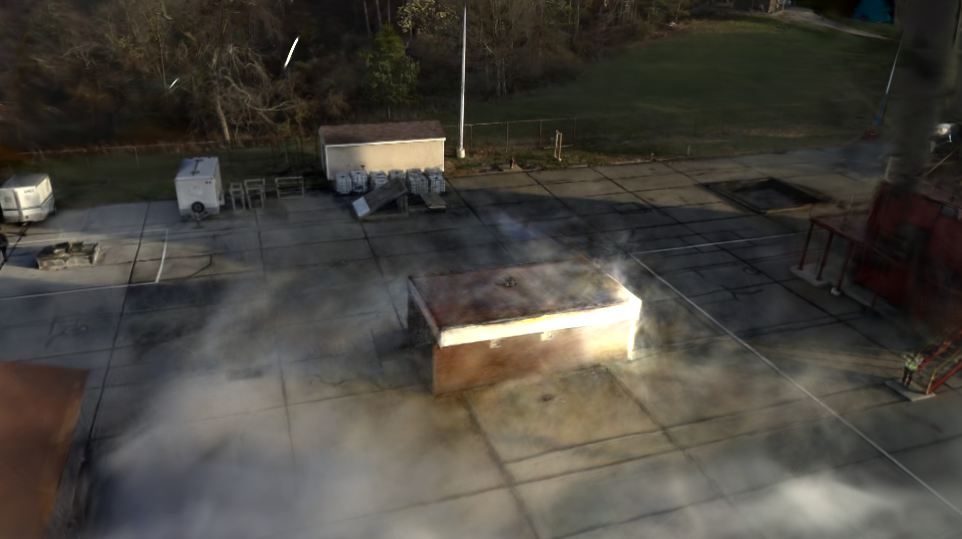} &
      \includegraphics[width=0.165\linewidth]{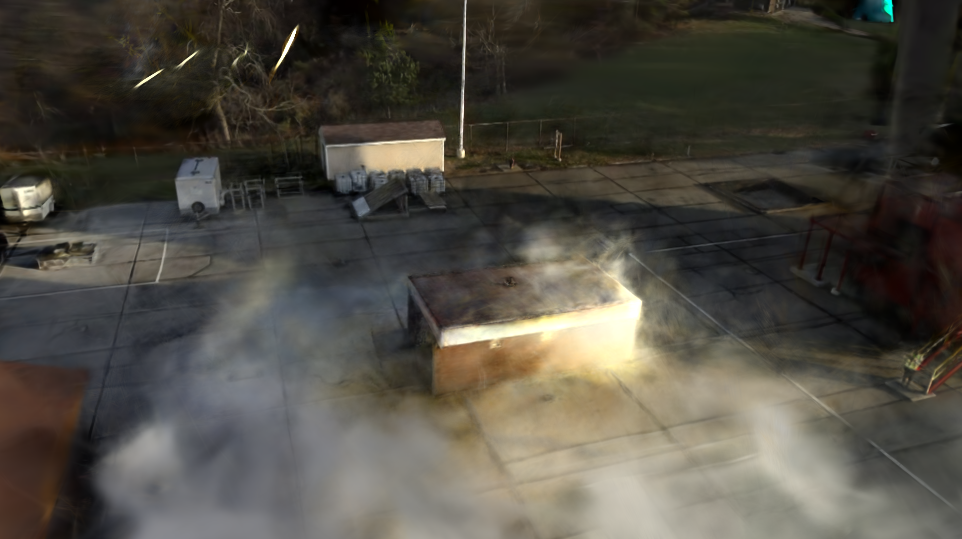} &
      \includegraphics[width=0.165\linewidth]{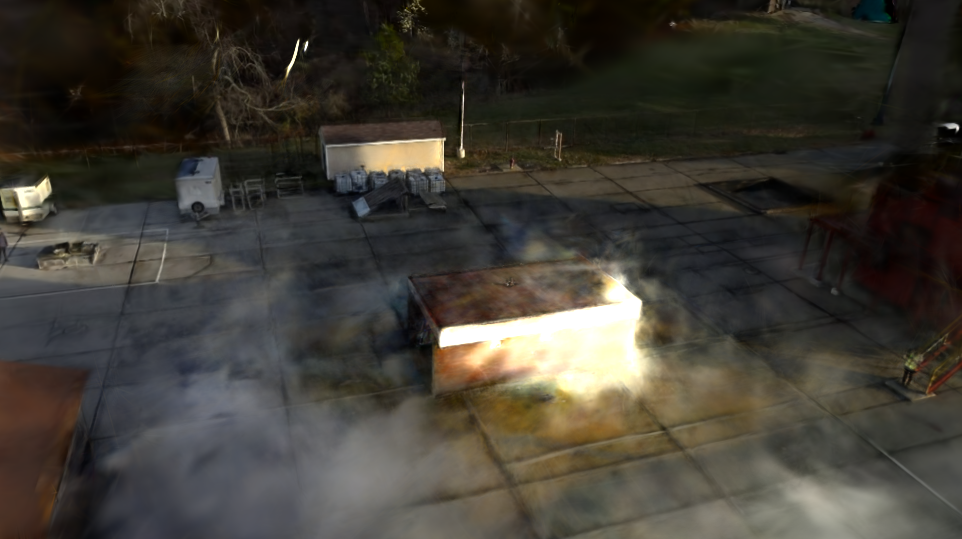} &
      \includegraphics[width=0.165\linewidth]{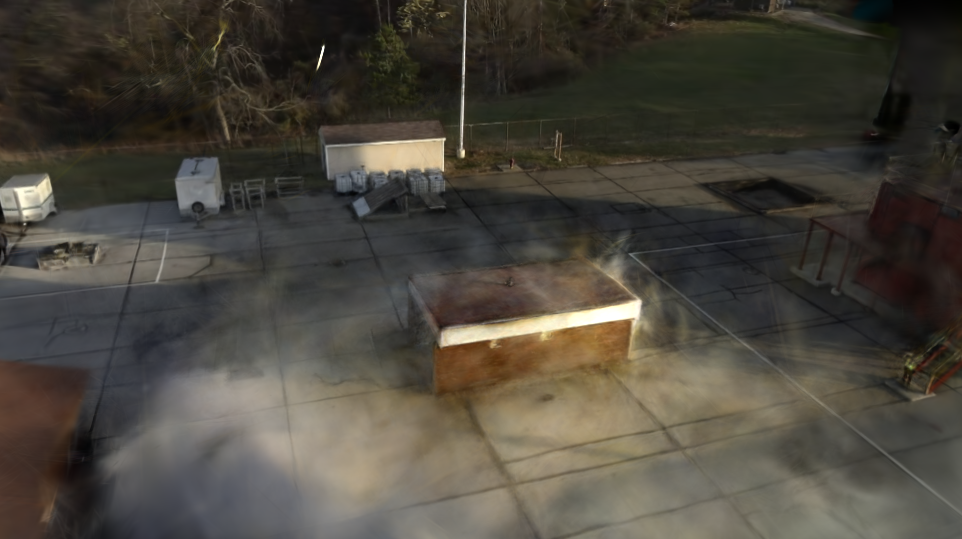}\\[-0.5ex]
  \end{tabular}
  \caption{Visual comparison of ablation configurations. Each image shows the result of removing a specific prior from our full method. Our method is able to recover the bricks in the wall of the house better than other configurations.}
  \label{fig:ablation}
\end{figure*}

% \subsection{Implementation details}
% \label{subsec:implementation}

% We train our models using the Adam optimizer \cite{Adam}. 
% We use the same hyperparameters for both synthetic and real-world experiments. We train Stage 2 for 15000 iterations and Stage 3 for 15000 iterations.
% For a typical scene in our real-world dataset, our method takes 1 hour for Stage 1, 
% 10 minutes for Stage 2 and 30 minutes for Stage 3.
\subsection{Baseline methods}
\label{subsec:baselines}

We compare three methods:

\begin{itemize}
    \item {\it ImgDehaze + 3DGS}: A two-stage approach that first applies a state-of-the-art single-image dehazing method (ConvIR \cite{cui2024revitalizing}) to each RGB frame, then uses deformable 3DGS \cite{yang2023deformable3dgs} on the dehazed images.
    \item {\it Ours (RGB only)}: Our approach using only RGB images (Stage 3 without thermal input).
    \item {\it Ours (Full)}: Our complete approach using both RGB and thermal images.
\end{itemize}
We could not compare with DehazeNeRF \cite{Chen2024dehazenerf}, the prior work closest to ours, due to lack of open-source code.

\subsection{Results}
\label{subsec:results}

\begin{table}
    \centering    
    \begin{tabular}{lccc}
      \toprule
      Method & PSNR $\uparrow$ & SSIM $\uparrow$ & LPIPS $\downarrow$ \\
      \midrule
      ImgDehaze + 3DGS & 14.42 & 0.37 & 0.318 \\
      Ours (RGB only) & 15.08 & 0.38 & 0.326 \\
      Ours (Full) & \textbf{19.92} & \textbf{0.76} & \textbf{0.247} \\
      \bottomrule
    \end{tabular}
    \caption{Quantitative results on the synthetic dataset for novel view synthesis. 
    Our full method outperforms all baselines.}
    \label{tab:synthetic_quantitative}
\end{table}

\paragraph{Synthetic data results.}
Table \ref{tab:synthetic_quantitative} presents quantitative results for novel view synthesis on our synthetic dataset, using the PSNR, SSIM, and LPIPS \cite{zhang2018perceptual} metrics. Our full method outperforms all baselines across all metrics, especially in scenes with heavy smoke where we achieve a PSNR gain of up to 4.8 dB over the RGB-only approaches.

Figure~\ref{fig:synthetic_qualitative} shows qualitative results on the synthetic dataset. Our method removes smoke while preserving fine details in the scene. In contrast, baseline methods either fail to completely remove smoke or introduce artifacts.

\vspace{-0.5cm}
\paragraph{Real-world data results.}
Figure~\ref{fig:real_qualitative} demonstrates our method's effectiveness on real-world data. 
The improvement is particularly noticeable in regions with dense smoke, where baseline 
methods struggle to reconstruct the scene geometry. Our method fuses complementary information in RGB and thermal modalities, 
to improve smoke removal while preserving texture details critical for scene understanding.

Though real-world reconstructions exhibit some artifacts, e.g., residual wisps of smoke or 
reduced color saturation under extreme occlusion (Figure~\ref{fig:real_qualitative}), 
they represent a substantial improvement in situational awareness. 
For first responders, the ability to discern room layout, locate doorways, and identify 
obstacles even at reduced fidelity turns an unusable, smoke-obscured video stream 
into an actionable 3D map. 

\vspace{-0.2cm}
\subsection{Ablation study}
\label{subsec:ablation}

\begin{table}
    \centering
    \begin{tabular}{lccc}
      \toprule
      Configuration & PSNR $\uparrow$ & SSIM $\uparrow$ & LPIPS $\downarrow$ \\
      \midrule
      w/o $\mathcal{L}_{\mathrm{smoke\_alpha}}$ & 18.96 & 0.68 & 0.312 \\
      w/o $\mathcal{L}_{\mathrm{smoke\_color}}$ & 19.02 & 0.71 & 0.289 \\
      w/o $\mathcal{L}_{\mathrm{mono}}$ & 19.78 & 0.76 & 0.248 \\
      w/o $\mathcal{L}_{\mathrm{depth}}$ & 19.88 & 0.75 & 0.252 \\
      w/o $\mathcal{L}_{\mathrm{mask}}$ & 19.82& 0.74 & 0.251 \\
      Ours (Full) & \textbf{19.92} & \textbf{0.76} & \textbf{0.247} \\
      \bottomrule
    \end{tabular}
    \caption{Ablation study showing the impact of each component in our framework. Each prior contributes to the overall performance, with the depth consistency prior having the most significant impact.}
    \label{tab:component_ablation}
\end{table}

Table \ref{tab:component_ablation} shows an ablation study on individual components in our framework. Each component provides a measurable performance improvement, and the combination of all components yields the best results. 
The depth consistency prior ($\mathcal{L}_{\mathrm{depth}}$) significantly improves performance, 
highlighting the importance of leveraging thermal information for accurate geometry reconstruction in smoke-filled environments.

Figure~\ref{fig:ablation} provides a visual comparison of different ablation configurations. 
Without the smoke consistency priors ($\mathcal{L}_{\mathrm{smoke\_alpha}}$ and $\mathcal{L}_{\mathrm{smoke\_color}}$), 
the model struggles to separate smoke from surfaces. Without the monochromaticity prior ($\mathcal{L}_{\mathrm{mono}}$), 
the model generates unrealistic colored smoke. The depth consistency prior ($\mathcal{L}_{\mathrm{depth}}$) 
is important for real-world scenes where the camera poses might be noisy. The mask alignment prior ($\mathcal{L}_{\mathrm{mask}}$) 
helps localize smoke and place smoke Gaussians in the correct location.

\vspace{-0.3cm}
\section{Conclusion}

We presented SmokeSeer, a framework for joint 3D scene reconstruction and smoke removal in dynamic smoke-filled environments. 
Our key insight is to leverage the complementary strengths of RGB and thermal imaging 
to decompose the scene into its surface and smoke components. 
We achieved this using a 3DGS-based inverse rendering pipeline to 
optimize separate smoke and surface Gaussians, appropriately 
regularized to account for their different physical properties. We demonstrated our method on synthetic datasets 
and real-world environments representative of firefighting settings. 
Our experiments in real-world firefighting scenarios demonstrate practical viability for emergency response applications. By publicly 
releasing our code and dataset, we aim to establish a foundation for future research in vision through smoke and multimodal scene understanding.

\vspace{-0.3cm}
\paragraph{Limitations and future work.}
Though our method achieves state-of-the-art performance in smoke removal and 3D reconstruction, 
several limitations remain. First, we model the temporal evolution of smoke using a deformation field, 
without explicitly incorporating physics-based priors such as fluid dynamics. 
Future work could integrate priors based on the Navier-Stokes equations to better capture smoke's physical behavior \cite{pinn}. Second, our method requires careful balancing of multiple loss terms during optimization (Equation \ref{eq:loss}), 
and cannot handle very dense smoke that might occlude the scene completely. 
Future work could incorporate generative priors to provide stronger guidance 
at regions heavily occluded by smoke. 

\vspace{-0.3cm}
\paragraph{Acknowledgments.} We thank Ian Higgins and John Keller for help with flying the drone during data acquisition; Sreekar Ranganathan for help with the camera hardware and data collection setup; and Jeff Tan and Nikhil Keetha for helpful discussions. This work was supported by National Institute of Food and Agriculture award 2023-67021-39073; Defense Science and Technology Agency contract \#DST000EC124000205; and Alfred P. Sloan Research Fellowship FG202013153 for Ioannis Gkioulekas. This work used Bridges-2 at the Pittsburgh Supercomputing Center, through allocation cis220039p from the Advanced Cyberinfrastructure Coordination Ecosystem: Services \& Support (ACCESS) program, which is supported by National Science Foundation awards 2138259, 2138286, 2138307, 2137603, and 2138296; and National Artificial Intelligence Research Resource Pilot-2211.
{
    \small
    \bibliographystyle{ieeenat_fullname}
    \bibliography{main}
}
\maketitlesupplementary
This supplement provides additional visualizations and implementation details that complement the main paper. 

\section{Smoke Segmentation Results}
\label{sec:supp_masks}
\begin{figure}[!t]
    \centering
    \begin{tabular}{cc}
        \toprule
        \textbf{Input RGB} & \textbf{RGB with Smoke Mask} \\
        \midrule
        \includegraphics[width=0.48\linewidth]{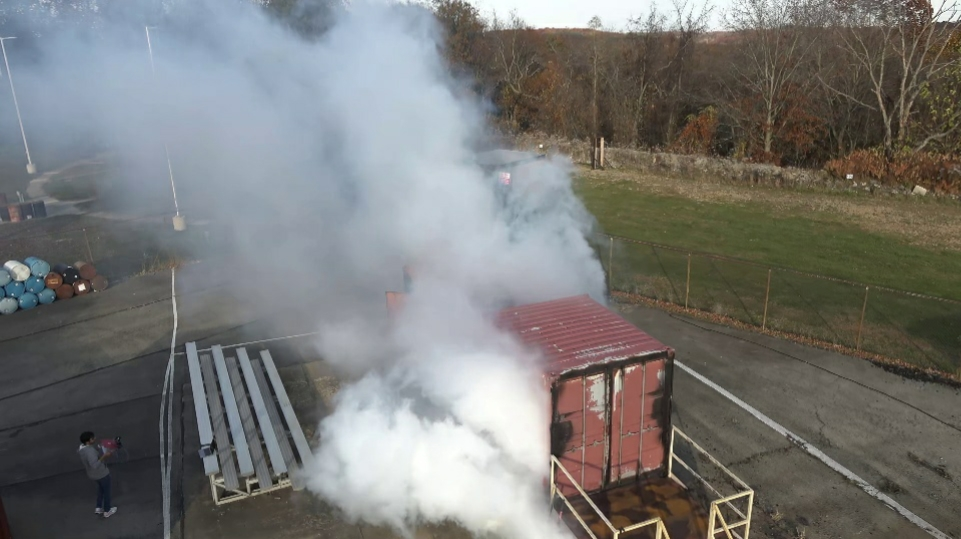} & \includegraphics[width=0.48\linewidth]{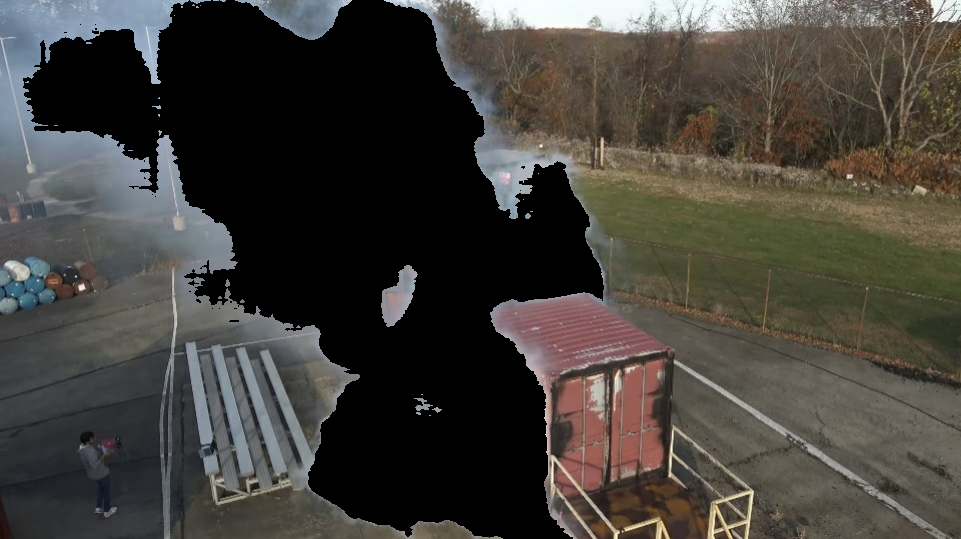} \\
        \includegraphics[width=0.48\linewidth]{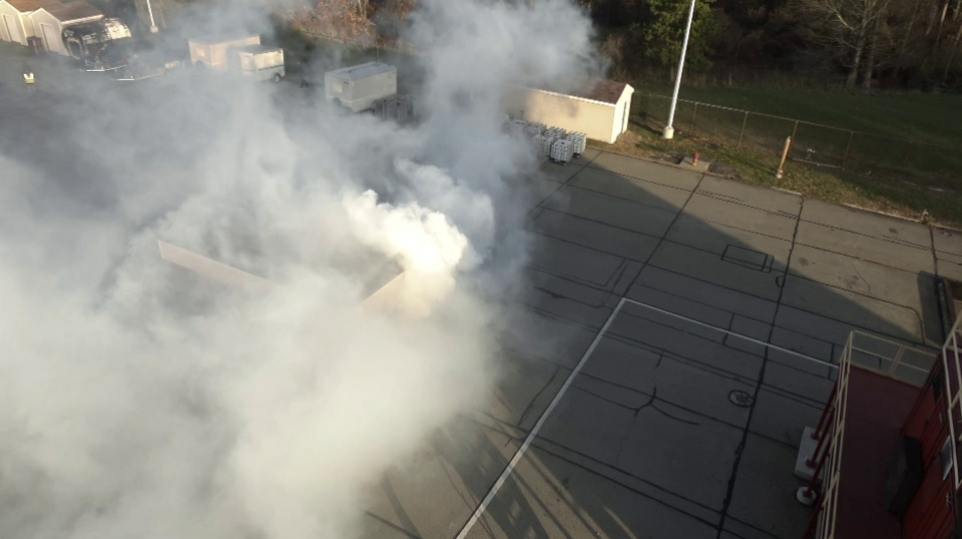} & \includegraphics[width=0.48\linewidth]{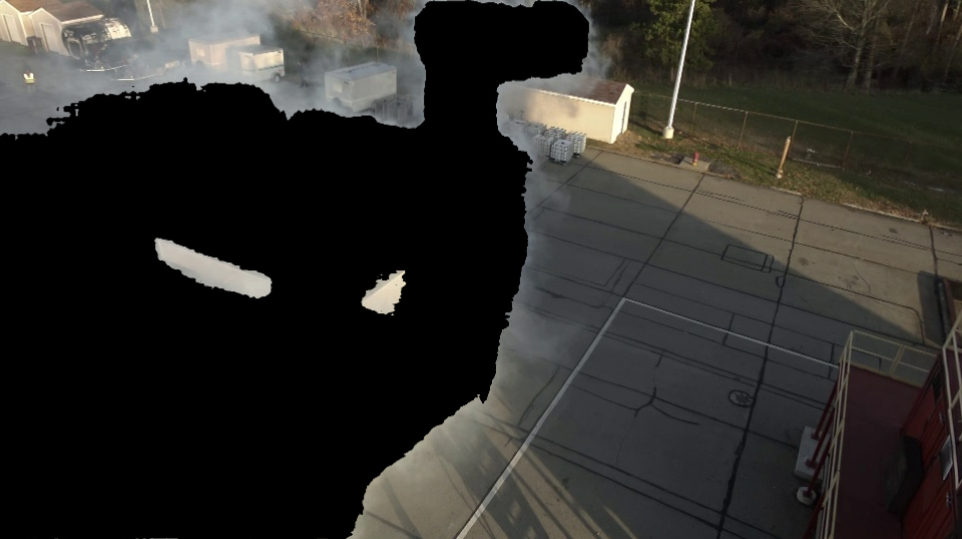} \\
        \bottomrule
    \end{tabular}
	\caption{Output of smoke segmentation pipeline. From left to right: Input RGB images with smoke, generated smoke masks. Note how the segmentation model accurately identifies smoke regions even with varying density and illumination.}
    \label{fig:smoke_masks}
\end{figure}

Grounded-SAM \cite{ren2024grounded} integrates Grounding DINO \cite{liu2023grounding}, an open-set object detector, with the Segment Anything Model (SAMv2) \cite{ravi2024sam2segmentimages}, to facilitate text-driven object detection and segmentation. We use this framework to automatically generate segmentation masks for smoke by inputting the prompt ``smoke.'' These masks are crucial for reliable feature matching in Stage 1 of our 
pipeline, and also serve as supervision for the mask alignment loss in Stage 3. Figure~\ref{fig:smoke_masks} shows example smoke segmentation masks.

\section{Feature Matching Comparison}
\label{sec:supp_feature_matching}

\begin{figure}[!t]
    \centering
    \includegraphics[width=\linewidth]{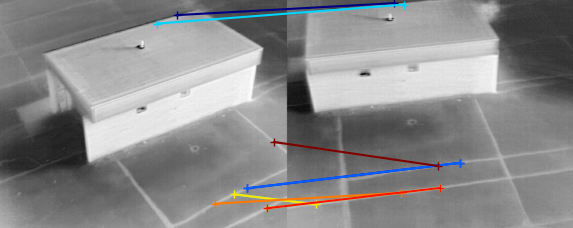} \\
    \includegraphics[width=\linewidth]{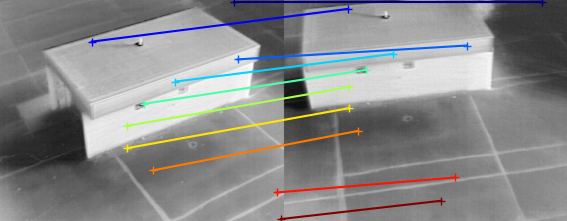}
    \caption{Feature matching comparison between images of the same modality (thermal). Top: SIFT feature matching fails in low-texture, low-contrast thermal images. Bottom: MAST3R-SfM provides more reliable correspondences by leveraging learned features that are more robust to the challenges of thermal imagery.}
    \label{fig:feature_matching}
\end{figure}

Figure~\ref{fig:feature_matching} compares traditional SIFT matching \cite{lowe1999object} with MAST3R-SfM~\cite{Duisterhof2024MASt3RSfMAF} 
for feature matching in smoke-affected scenes. The comparison highlights how SIFT matching 
fails for low-texture thermal images, while MAST3R-SfM with smoke masking provides more reliable correspondences. This robust matching is essential for 
the accurate camera pose estimation required in Stage 1 of our pipeline.

\section{Cross-Modal Registration Details}
\label{sec:supp_cross_modal}

\begin{figure}[!t]
    \centering
    \includegraphics[width=\linewidth]{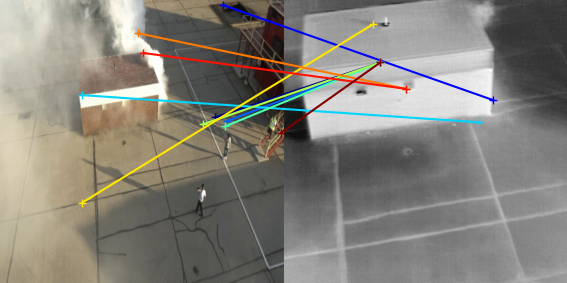} \\
    \includegraphics[width=\linewidth]{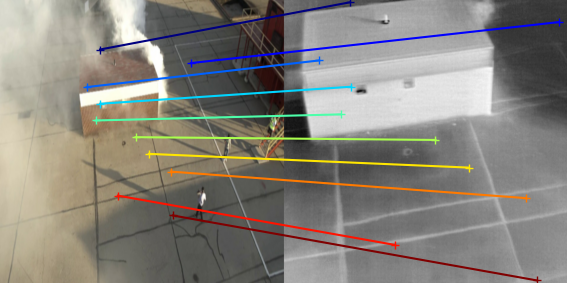}
    \caption{Cross-modal feature matching comparison. Top: Traditional SIFT feature matching fails 
    between RGB and thermal images due to fundamental differences in appearance across modalities. 
    Bottom: MINIMA \cite{jiang2025minima} provides more reliable correspondences 
    by explicitly addressing cross-modal challenges, enabling accurate registration of the two sensor types.}
    \label{fig:cross_modal}
\end{figure}

Figure~\ref{fig:cross_modal} visualizes the cross-modal registration process using MINIMA~\cite{jiang2025minima} 
for aligning RGB and thermal coordinate systems. We find that running COLMAP~\cite{schoenberger2016sfm} 
on our data fails to register the images and gives a degenerate result. Running MAST3R-SfM~\cite{Duisterhof2024MASt3RSfMAF} 
does give better results than COLMAP and is able to register all images. However, it is still not perfect and there is some misalignment.
The visualization shows the 2D correspondences established between RGB and 
thermal image pairs and the resulting aligned point clouds. This cross-modal registration 
is crucial for our approach, as it allows us to get the RGB and thermal images in the same coordinate system which is necessary for running
the subsequent 3D reconstruction pipeline.

\section{Implementation Details}
\label{sec:supp_implementation}

We provide implementation details to facilitate reproducibility. We implemented our framework in PyTorch, 
building on the official 3DGS repository. 
For deformable Gaussian splatting, we adapted the implementation from Yang et al.~\cite{yang2023deformable3dgs}.
We used the default parameters and configurations for 3DGS and Deformable 3DGS. We used the same hyperparameters for both synthetic and real-world experiments. We trained Stages 2 and 3 for 15000 iterations each, using the Adam optimizer \cite{Adam}. 
For Stage 3, we set $\lambda_{\mathrm{render}} = 1.0$, $\lambda_{\mathrm{smoke\_alpha}} = 0.1$, $\lambda_{\mathrm{smoke\_color}} = 0.05$, $\lambda_{\mathrm{mono}} = 0.1$, $\lambda_{\mathrm{depth}} = 2.0$, and $\lambda_{\mathrm{mask}} = 0.5$.
We ran all experiments on a workstation with an NVIDIA RTX 4090 GPU, an Intel i9-13900K CPU, and 128 GB RAM. For a typical scene in our real-world dataset, our method takes 1 hour for Stage 1, 10 minutes for Stage 2 and 30 minutes for Stage 3.

\section{Reference for real-world experiments}
We do not report quantitative metrics for the real-world dataset as obtaining true ground truth is challenging in such environments.
Instead, we provide an approximation that we refer to as ``Reference'' in figures. To create this reference, we collected additional smoke-free RGB images of the same scenes in a separate drone flight. We then: (1) reconstructed the smoke-free scene using 3DGS,
(2) obtained the camera poses of smoke-filled and smoke-free image sets in the same coordinate frame, and
(3) rendered novel views of the smoke-free reconstruction using the camera poses from the smoke-filled sequence.
These reference images serve as an approximate benchmark, though they are not perfect ground truth as the poses are noisy and environmental conditions may have changed between captures.

\end{document}